\documentclass{article}

\usepackage[english]{babel}
\usepackage{amssymb} 
\usepackage[letterpaper,top=2cm,bottom=2cm,left=3cm,right=3cm,marginparwidth=1.75cm]{geometry}

\usepackage{placeins}
\usepackage{amsmath}
\usepackage{authblk}
\usepackage{natbib}
\usepackage{graphicx}
\usepackage{tabularx}
\usepackage{makecell}
\usepackage{subcaption}
\usepackage{appendix}
\usepackage{abstract}
\usepackage{comment}
\usepackage[colorlinks=true, allcolors=blue]{hyperref}
\usepackage{float}
\usepackage{longtable}
\usepackage{booktabs}
\usepackage{siunitx}

\title{Robust Route Planning for Sidewalk Delivery Robots}
\author[a]{Xing Tong}
\author[a]{Michele D. Simoni\thanks{Corresponding author. Email: micheles@kth.se}}

\affil[a]{Division of Transport and Systems Analysis, KTH Royal Institute of Technology, Teknikringen 10A, Stockholm 10044, Sweden}

\begin{document}
\maketitle

\begin{abstract}
Sidewalk delivery robots are a promising solution for last-mile freight distribution. Yet, they operate in dynamic environments characterized by pedestrian flows and potential obstacles, which make travel times highly uncertain and can significantly affect their efficiency. This study addresses the robust route planning problem for sidewalk robots by explicitly accounting for travel time uncertainty generated through simulated interactions between robots, pedestrians, and obstacles. Robust optimization is integrated with simulation to reproduce the effect of obstacles and pedestrian flows and generate realistic travel times. Three different approaches to derive uncertainty sets are investigated, including budgeted, ellipsoidal, and support vector clustering (SVC)-based methods, together with a distributionally robust shortest path (DRSP) method based on ambiguity sets that model uncertainty in travel-time distributions. A realistic case study reproducing pedestrian patterns in Stockholm’s city center is used to evaluate the efficiency of robust routing across various robot designs and environmental conditions. Results show that, when compared to a conventional shortest path (SP) method, robust routing significantly enhances operational reliability under variable sidewalk conditions. The ellipsoidal and DRSP approaches outperform the other methods in terms of average and worst-case delay. Sensitivity analyses reveal that robust approaches are higher for sidewalk delivery robots that are wider, slower, and more conservative in their navigation behaviors, especially in adverse weather and high pedestrian congestion scenarios. 
\end{abstract}

\section{Introduction}

The last-mile delivery problem represents one of the most challenging aspects of modern logistics due to its cost, increased traffic congestion, and environmental impacts, particularly in urban areas \citep{boysen2021last}. The growing demand for e-commerce and the need for faster and more reliable deliveries have led to the development of innovative solutions such as drones and autonomous ground robots \citep{Jennings2019}.

Sidewalk delivery robots, despite their lower speeds and reliance on existing infrastructure, represent a safer and higher-capacity solution than drones, making them more suitable for dense urban environments. In recent years, the robots have moved beyond small-scale pilot tests and have been increasingly deployed in real-world commercial services, particularly for food and parcel delivery. Major logistics and platform operators have introduced sidewalk robots for last-mile operations in multiple cities, operating from restaurants, dark kitchens, and local depots to serve customers in residential and commercial districts. For example, Serve Robotics has expanded its autonomous delivery fleet through partnerships with Uber Eats and DoorDash in several U.S. cities, while Starship Technologies has exceeded nine million autonomous deliveries worldwide across food and grocery sectors \citep{Serve2025,Starship2025}. In parallel, DoorDash and Uber Eats have launched autonomous sidewalk delivery services in large metropolitan areas such as Los Angeles, Miami, and Jersey City \citep{UberEats2025,DoorDash2025}. Beyond stand-alone robot operations, hybrid delivery concepts combining trucks and robots have also been explored, where robots are deployed from the back of the vans to reach the customer \citep{starshipMercedes}. However, such integrated vehicle–robot delivery systems remain relatively uncommon in current commercial practice. These developments demonstrate that sidewalk robot delivery is transitioning from experimental trials to scalable urban logistics solutions, increasing the importance of reliable and robust operational planning methods.

While sidewalk delivery robots operate at relatively low speeds, their efficiency can be significantly affected by factors like pedestrian traffic, terrain, road conditions, obstacles, and weather \citep{HEIMFARTH2022401}. These external variables introduce substantial travel time uncertainty, a challenge that has not been thoroughly addressed in previous studies on robot-based delivery. Existing approaches relying on static travel times for sidewalks may lead to suboptimal operational and strategic decisions (e.g., routing, assignment, location problems) by overlooking potential delays. 

In this study, we address robust path planning for sidewalk delivery robots by explicitly accounting for travel time variability. Rather than assuming parametric travel times, we integrate robust and distributionally robust methods with microscopic pedestrian simulation to endogenously derive travel times from interactions between robots, pedestrians, and obstacles. Specifically, pedestrian simulation enables the generation of realistic heterogeneous conditions at the detailed level. These input are then used to define multiple uncertainty representations required to solve the Robust Shortest Path Problem (RSPP) and Distributionally Robust Shortest Path Problem (DRSP). Two widely-adopted approaches, interval-budgeted and ellipsoidal, and an advanced data-driven method, the kernel-based support vector classification (SVC), are investigated to define uncertainty sets in the RSPP. To assess the generality of the results, we analyze a set of synthetic network instances characterized by different combinations of pedestrianf flows and obstacles. In addition, a realistic urban pedestrian simulation scenario, set in a central district of Stockholm, Sweden, is used to analyze the performance of alternative methods over conventional approaches. This area was designated as a Class 3 clean air zone, a policy aimed at drastically reducing urban vehicle emissions by permitting only fully electric vehicles, making it particularly relevant for studying sidewalk robot delivery \citep{cleanairzone}. In the second part of the paper, we conduct a comprehensive analysis of how design-related factors (e.g., robot maximum speed, size, navigation behavior) and environmental factors (e.g., pedestrian flows, weather) affect the efficiency of robust versus conventional routing.

The main contributions of this study are threefold. First, we investigate the robust route planning problem for sidewalk robot navigation and provide a framework for uncertainty-aware routing that is applied to both synthetic and realistic sidewalk networks. Within this framework, travel time uncertainty is generated endogenously with simulation from the interactions among robots, obstacles, and pedestrians, generating realistic travel times. 
Second, we investigate two alternative data-driven approaches to derive uncertainty sets in the RSPP, such as the kernel-based SVC and the distributionally robust shortest path, which minimize assumptions and parameter tuning to endogenously balance conservatism and efficiency.
These approaches are systematically evaluated against more standard approaches based on budgeted and ellipsoidal uncertainty sets, to highlight their relative strengths and weaknesses under the comprehensive evaluation framework we proposed for robust routing methods. Third, we derive operational insights by analyzing the sensitivity of robust routing performance through a systematic analysis of critical factors, such as robot speed, size, maneuverability, pedestrian flows, and weather conditions.

The remainder of this paper is structured as follows. Since the research fields of robot operational planning and robust route planning are distinct, we briefly review the relevant literature for each in Section \ref{Sec:Related Studies}. Section \ref{Sec:Methodological Approach} introduces the methodological approach, covering the four different robust optimization approaches and sidewalk robot simulation. Section \ref{Sec:Results} provides case studies on both a synthetic network and a realistic network in central Stockholm, and evaluates the efficiency of robust optimization approaches under various conditions through sensitivity analyses. Section \ref{Sec:Conclusion} concludes the study by summarizing key insights and outlining avenues for future research.

\section{Related studies}
\label{Sec:Related Studies}

\subsection{Sidewalk autonomous delivery robots}

Sidewalk Autonomous Delivery Robots (SADRs) represent one of the three categories of Autonomous Delivery Robots (ADRs) employed for delivery applications, together with Road Autonomous Delivery Robots (RADRs) and Autonomous Delivery Vehicles (ADVs), as outlined by \citet{srinivas2022autonomous} who provide a comprehensive overview of robot-related studies. SADRs are pedestrian-sized robots designed to travel on sidewalks to deliver items autonomously.  In this review, the focus is specifically on SADRs, which are referred to as delivery robots for simplicity. 

\citet{Jennings2019} conducted an in-depth study on robots, discussing current regulations and their technical capabilities, and proposed a model to investigate their impact in truck-robot routing problems. Their findings suggest that adopting robots can significantly reduce delivery times, on-road vehicle travel, and costs in comparison to conventional deliveries, particularly in densely populated urban areas. Similar conclusions were drawn by \citet{lemardele2021potentialities} who estimated the operations costs and external impacts of delivery robots through case studies in Paris, France, and Barcelona, Spain. Furthermore, \citet{marks2019robots} conducted a detailed analysis of the regulatory framework around delivery robots in the US. 

The most common research focus for SADRs is the truck-robot routing problem, also named as the Traveling Salesman Problem with Robots (TSP-R), which leverages the bulk transportation capabilities of trucks with the flexibility and efficiency of small robots that handle the last mile of deliveries. While the problem can be considered as a particular case of the Traveling Salesman Problem with Drones (TSP-D), the use cases for robots differ significantly. From an operational perspective, robots have notably slower speeds (5–10 km/h compared to 50–100 km/h) and can travel shorter distances (5–10 km compared to 10–30 km).  These differences make robots particularly well-suited to short-range delivery of low-value items in dense urban environments, where operating at pedestrian speed and carrying higher payloads than drones can improve safety and practicality \citep{SIMONI2020102049}. Different ways to combine trucks and robots in delivery have been discussed by many researchers. \citet{BOYSEN20181085} explored scheduling models specifically for truck-based robot delivery. Their research highlights the potential benefits of using trucks as launching platforms for robots, which can then autonomously deliver goods to customers and return to centralized depots. \citet{liu2021hybrid} developed a two-tier model in which vehicles transfer parcels from depots to satellites in the first tier and robots deliver parcels from satellites to customers in the second tier. \citet{CHEN2021102214} investigated a vehicle routing problem with time windows and delivery robots (VRPTWDR), which focuses on the benefits of deliveries made by both trucks and robots. Their work considers the case that customers need to be present to pick up the parcels when the robots arrive.  However, their models are limited by sequential delivery actions of trucks and robots, which means trucks and robots can not make deliveries in parallel. Building on this, \citet{HEIMFARTH2022401} proposed a mixed truck-robot delivery system, introducing additional flexibility by allowing trucks to deliver larger or sensitive packages, while robots handle smaller parcels. Their mixed truck-robot routing problem demonstrated significant cost reductions compared to traditional truck-only or truck-and-robot systems without simultaneous deliveries by both truck and robots within the same tour. Similarly, \citet{MURRAY201586} investigated synchronized truck-UAV deliveries, where the UAV is launched from and retrieved by the truck. This formulation laid the foundation for a growing body of research on hybrid truck–drone routing. 

Several studies have developed frameworks for robot-only delivery, where parcels are transported exclusively by a fleet of robots without relying on vehicles like trucks or vans. However, these frameworks often encompass various types of delivery robots, not limited to sidewalk robots. \citet{ulmer2019same} considered a scenario in which ADRs are dispatched directly from depots to pickup stations, and analyzed the operational level of same-day delivery with pickup stations and ADRs. \citet{li2021deploying} proposed a collaboration framework between ADRs and couriers, where the ADRs transfer parcels between the couriers and the depot, and thus, the couriers can keep making deliveries without returning to the depot. Extending beyond traditional ADRs, \citet{ENSAFIAN2023103958} investigated a two-echelon network involving autonomous mobile lockers (AMLs) working in coordination with couriers. Their model allows open routes and multi-trip missions, improving operational flexibility and reducing delivery costs. Beyond human–robot collaboration, multimodal delivery models have also been explored. \citet{DEMAIO2024104615} formulated a routing problem in which ADRs autonomously synchronize with scheduled public transport lines to reach otherwise inaccessible customers, demonstrating significant cost and emission savings.  

The operation of sidewalk robots may lead to potential conflicts with pedestrians, as they must share the same sidewalk space. \citet{GEHRKE2023100789} observed that SADRs prompted pedestrians to alter their paths to avoid collisions, with the help of field-recorded video data from a university campus. Additionally, SADRs spend 30\% of their transit time waiting for pedestrians according to the study of \citet{Jennings2019}. These interactions between robots and pedestrians were taken into account in the sidewalk robot navigation approach proposed by  \citet{Du2019}.

Despite these advances, existing research assumes static or simplified travel conditions for the sidewalk robot routing problems, overlooking the impact of variable sidewalk travel times due to congestion and other factors. Robots, which typically travel at pedestrian speeds, may face significant delays when navigating crowded or obstructed sidewalks. The traditional shortest path considering length or fixed travel time may not perform effectively in all scenarios, especially when extreme conditions occur. This issue remains insufficiently addressed in the current sidewalk network robot routing problems. This paper addresses this gap by incorporating variable travel times for sidewalk robot delivery in congested areas, providing a more realistic and robust approach for robot-based last-mile delivery operations.

\subsection{Robust and distributionally robust shortest path problem}

The shortest path problem (SPP) is a fundamental problem in combinatorial optimization and graph theory that aims to find the most efficient route between two (or more) nodes in a network. It has broad applications in various fields, including transportation, planning, and network design \citep{Kumawat}. However, real-world network conditions are often uncertain due to fluctuating travel times, changing road conditions, or unpredictable delays. To address these uncertainties, the robust shortest path problem (RSPP) extends the classical SPP by incorporating uncertainty into the model, ensuring that the chosen path remains acceptable performance in terms of travel time even under adverse conditions. 

Robust Optimization (RO) approaches typically handle uncertainty by defining uncertainty sets that capture possible variations in arc costs or travel times. Commonly used types of uncertainty sets include convex hull \citep{Kasperski2016}, intervals \citep{CHASSEIN2015739}, ellipsoid \citep{BENTAL1998}, budgeted uncertainty \citep{Goerigk2016} and permutohull \citep{Bertsimas2009}. Each of these sets reflects different assumptions about the nature of uncertainty and requires distinct optimization techniques to solve the RSPP efficiently. These different types of uncertainty sets were tested by \citet{dokka2017} using real-world traffic measurements, and compared their performance under different performance indicators. \citet{chassein2018} built and compared a range of uncertainty sets, and developed an efficient solution algorithm focusing on the case of ellipsoidal uncertainty for RSPP. Different studies extended to solve more complex problems, such as the RSPP with time windows \citep{Alves2015}, and to explicitly account for accident probability and the accident consequence \citep{Kwon2013RobustSP}.

In recent years, data-driven approaches have moved beyond traditional uncertainty modeling by incorporating empirical insights, allowing for the dynamic adjustment to better align with observed variability and real-world trends. A growing body of research has investigated the distributionally robust shortest path problem (DRSP), where an ambiguity set (i.e., a collection of plausible cost distributions) is constructed based on the empirical distribution derived from data \citep{CHENG2013511,wang2019,KETKOV2021105212}. Although DRSP provides theoretical robustness guarantees, it is often constrained by the limited expressiveness of its uncertainty sets and can be computationally challenging for large-scale networks or high-dimensional settings. An alternative is to adopt machine learning techniques, which can better capture the complexity and asymmetry of real-world data. For example, \citet{SHANG2017464} proposed a Support Vector Clustering (SVC) to construct uncertainty sets, and later extended this method to handle high-dimensional uncertainty with decomposition techniques (\citep{SHANG201819}). These approaches can effectively handle data correlation and asymmetry, but since they are Single-Kernel Learning-based (SKL) methods in essence, their performance heavily depends on projection directions, and they may lead to over-conservative solutions, particularly for asymmetric data. \citet{HAN2021} proposed a Multiple-Kernel Learning-based (MKL) SVM approach that optimally selects projection directions to create non-conservative polyhedral uncertainty sets. However, this method is only suitable for low-dimensional uncertainty. \citet{GHIASVAND2023108390} developed a weighted one-class support vector machine (WOC-SVM) algorithm, in order to create adjustable uncertainty sets considering supplementary information, such as predicted values. \citet{GOERIGK2023} adopted deep neural networks to construct non-convex uncertainty sets, and integrated them into a robust optimization model by formulating the adversarial problem as a convex quadratic mixed-integer program. These advances demonstrate the potential of machine learning techniques, but still face computational challenges, especially for high-dimensional and large datasets. \citet{ROYTVANDGHIASVAND2024} addresses this problem by decreasing the number of support vectors through a novel hybrid two-stage clustering (TSC) algorithm. 

In this paper, we develop two data-driven methods to specifically address the robust routing planning in sidewalk robot delivery: a Wasserstein-based DRSP approach and a RSP method using SKL-based and MKL-based SVC approaches combined with the TSC algorithm. We evaluate and compare the performance of these methods against two established baselines through realistic simulation scenarios designed for sidewalk robot delivery. 

\section{Methodological approach}
\label{Sec:Methodological Approach}

In this study, we systematically evaluated the performance of four approaches: Budgeted uncertainty, Ellipsoidal uncertainty, Kernel-based SVC uncertainty, and Wasserstein-based DRSP ambiguity. These methods are assessed in the context of sidewalk robot delivery using pedestrian simulation data. 

Budgeted uncertainty, introduced by \citep{Bertsimas2003}, is a widely recognized and commonly adopted approach for RSPP, appearing extensively in both theoretical research and practical applications. It is computationally tractable and provides a tunable trade-off between robustness and conservatism. Ellipsoidal uncertainty typically provides stable solutions and performs well overall. It achieves the best trade-off compared to other approaches, including convex hull, intervals, budgeted, permutohull, and symmetric permutohull uncertainty, when applied to RSPP with real-world data \citet{chassein2018}.  

The first two approaches serve as traditional RSP methods. In contrast, the latter two are more recent data-driven methods. Kernel-based SVC uncertainty leverages machine learning to construct uncertainty sets from data but remains underexplored in the context of RSPP, especially for sidewalk robot delivery. We implement and evaluate two variants of this method: SKL-based SVC and MKL-based SVC. Additionally, we include the Wasserstein-based DRSP method, a prominent data-driven technique that offers strong theoretical guarantees and flexible ambiguity set construction based on empirical distributions. 

\begin{figure}[t]
    \centering
        \includegraphics[width=0.7\linewidth]{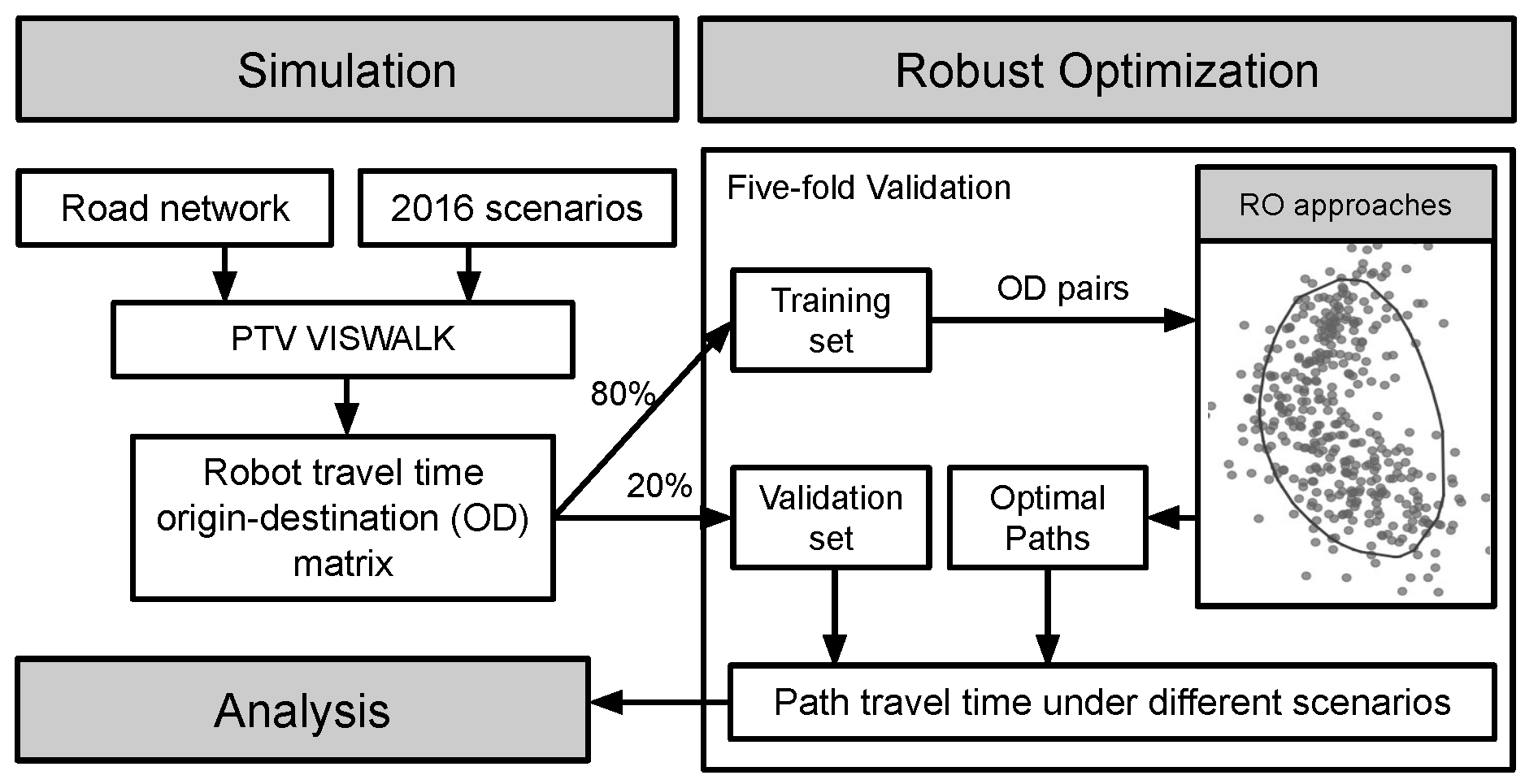}
    \caption{Framework for Simulation and Optimization}
    \label{fig:1}
\end{figure}

The framework of the integration of simulation and optimization is shown in Figure~\ref{fig:1}. An overview of the four approaches is provided in Section~\ref{sec:RRPA}. This is followed by the introduction of the model used to simulate the sidewalk robots in Section~\ref{sec:sidewalkrobot simulation}. A brief discussion on the implication for multi-stop tours is presented in Section \ref{sec:implications_multi_stop}. Table~\ref{tab:glossary} in Appendix~\ref{app:glossary} summarizes the main acronyms and notation used throughout the paper.

\subsection{Robust route planning approaches}
\label{sec:RRPA}
In the classic shortest path problem, we denote a directed graph as $G = (V,A)$ where $V$ indicates the set of vertices or nodes, and $A$ indicates the set of segments, \(|A|=n\). Each segment has a cost $c$. Given a start point $s$ and a target point $t$, the goal is to find a route with a minimum sum of costs between them. This problem can be defined briefly as follows:
\begin{equation}
\min \{\mathbf{c}^T \mathbf{x}:\mathbf{x} \in X\} 
\label{eq:1}
\end{equation}
where $X$ is the set of feasible paths defined by standard flow conservation and binary constraints from $s$ to $t$, and \( X \subseteq \{0,1\}^n \).   

In the robust shortest path problem, the segment costs \(c\) are not precisely known. There is a set of observations of costs \(D=\{\mathbf{c}^1,...,\mathbf{c}^N \}\), where \(N\) is the number of observations and \(\mathbf{c}^i \in \mathbb{R}^n \). We use the notation \([n] = \{1,2,...,n\}\) and \([N] = \{1,2,...,N\}\). Based on the set of observations \(D\), the uncertainty set \(U\), which contains plausible realizations of the cost vector can be modeled. The RSPP is then denoted as:
\begin{equation}
\min \{ \underset{\mathbf{u}\in U}{\max} \mathbf{u}^T \mathbf{x}:\mathbf{x} \in X\} \label{eq:2}
\end{equation}
which means finding the shortest path in \(X\)  that minimizes the worst-case cost over all realizations in \(U\), providing protection against adverse but bounded deviations in segment costs.

In contrast, the DRSP method does not fix a single probability distribution for costs, but assumes that the true distribution \(F\) belongs to an ambiguity set $\mathcal{F}$, which is a family of probability distributions consistent with the observed data. The corresponding DRSP problem then minimizes a worst-case risk measure (e.g., an expected or tail-based criterion) over all distributions in $\mathcal{F}$, thus hedging against errors in distributional assumptions rather than only against point-wise cost deviations. In data-driven settings, both $U$ and $\mathcal{F}$ are constructed from samples: uncertainty sets can be learned to approximate the geometric support of observed realizations (e.g., via kernel-based SVC), while ambiguity sets are typically defined as neighborhoods around the empirical distribution (e.g., Wasserstein balls), with their size controlling the degree of conservatism.

\begin{figure}[t]
    \centering
    \begin{subfigure}{0.48\textwidth}
        \centering
        \includegraphics[width=\linewidth]{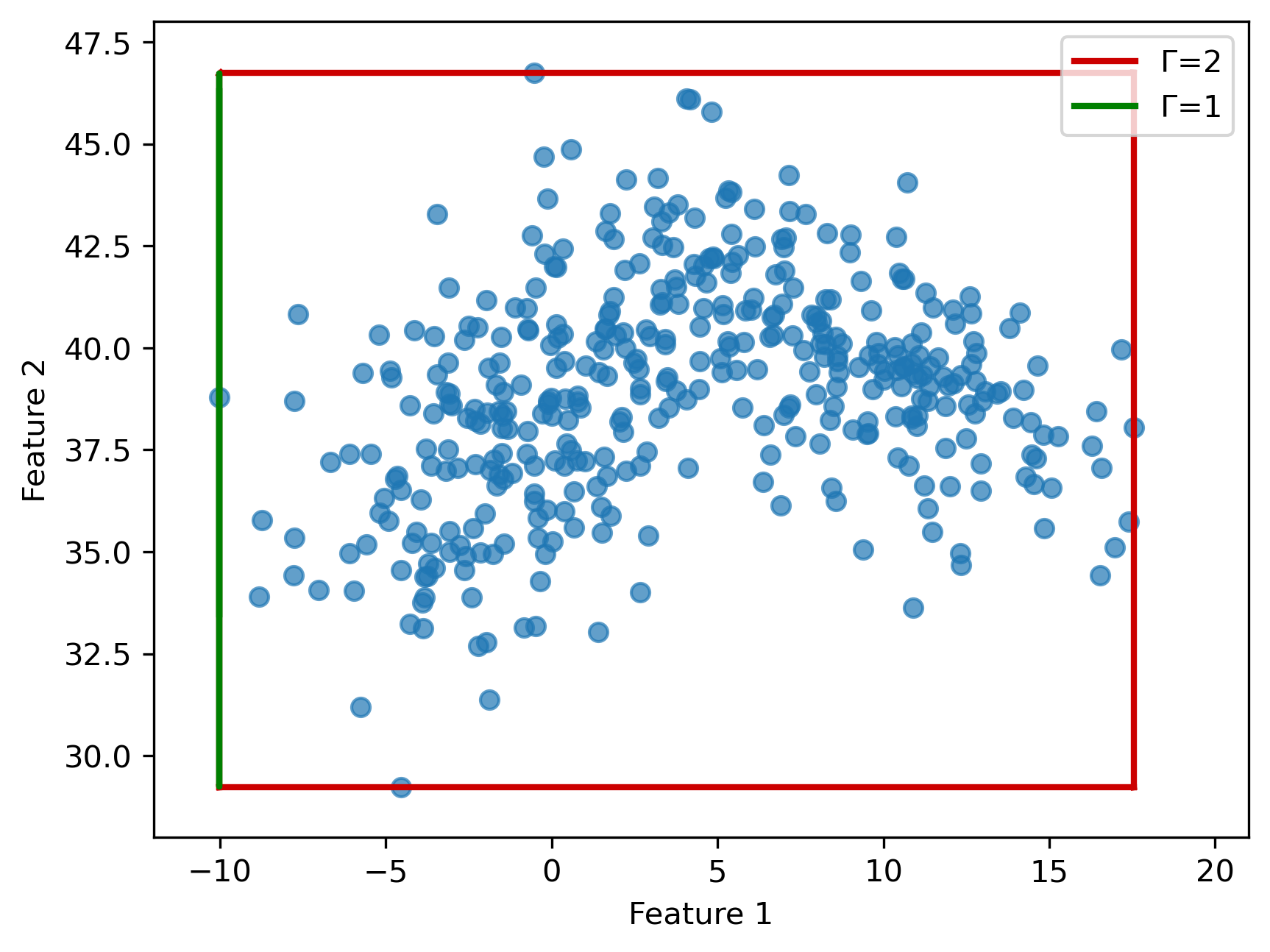}
        \caption{Budgeted uncertainty}
        \label{fig:2a}
    \end{subfigure}
    \hfill
    \begin{subfigure}{0.48\textwidth}
        \centering
        \includegraphics[width=\linewidth]{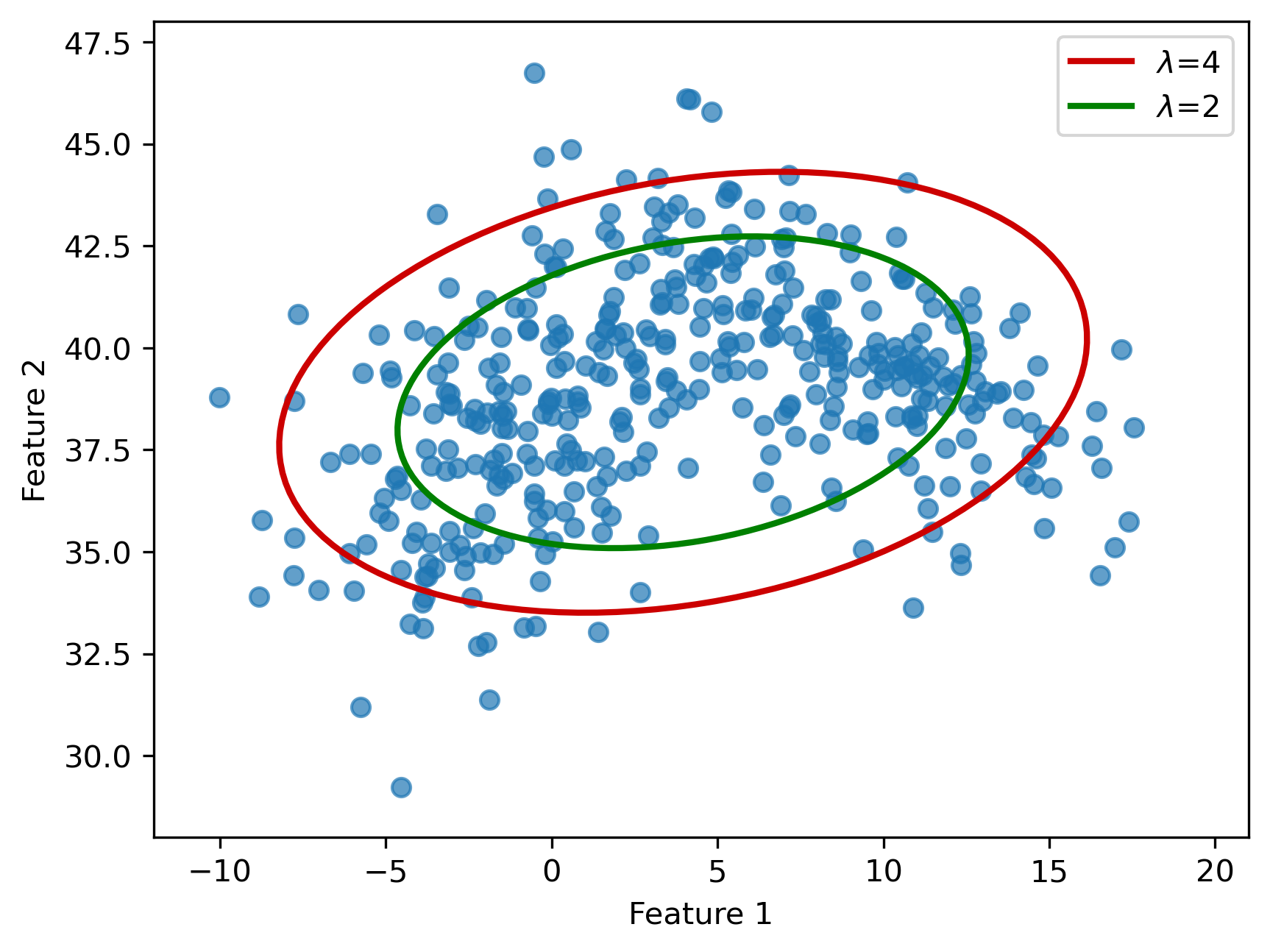}
        \caption{Ellipsoidal uncertainty}
        \label{fig:2b}
    \end{subfigure}
    \begin{subfigure}{0.48\textwidth}
        \centering
        \includegraphics[width=\textwidth]{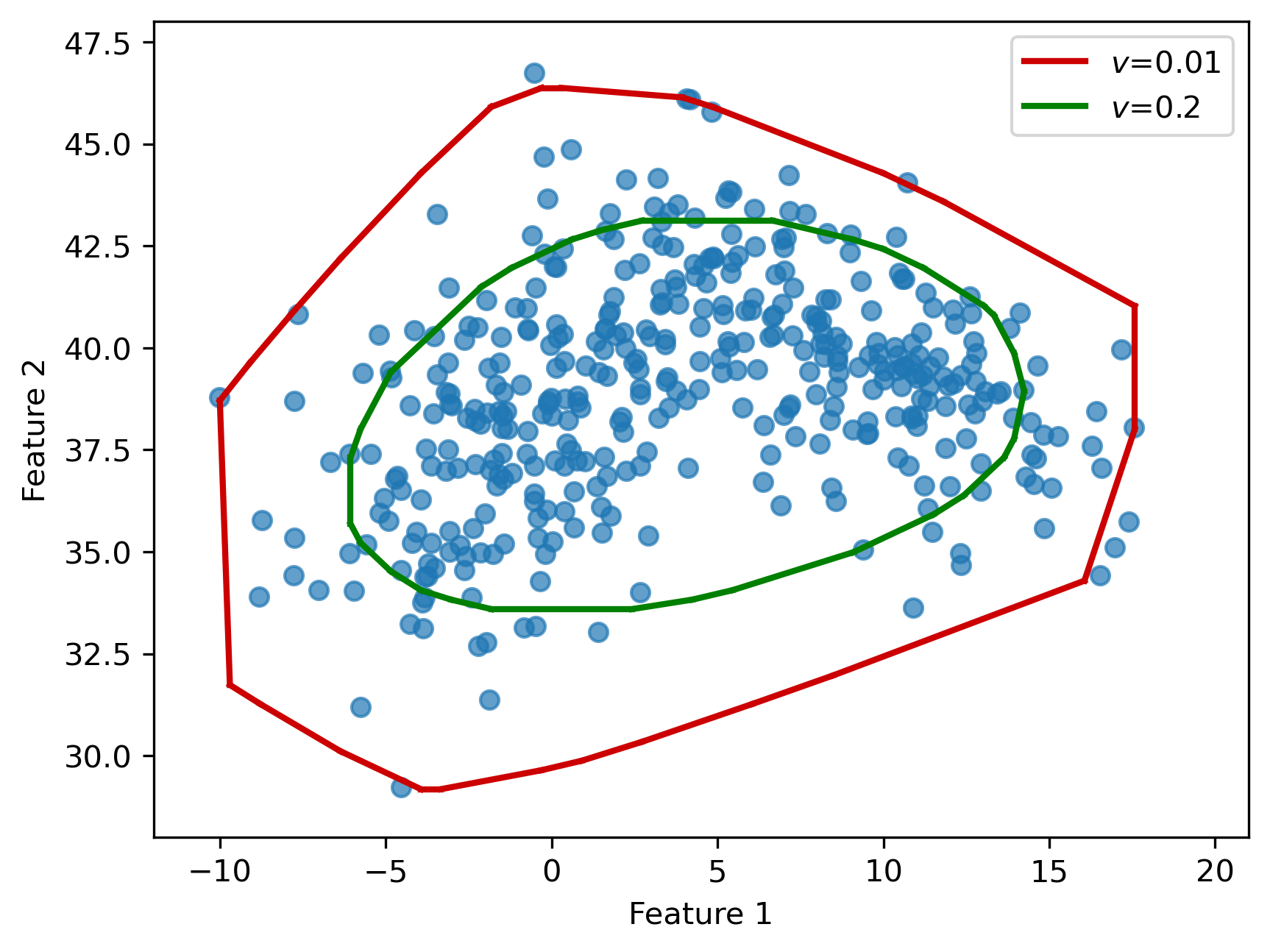} 
        \caption{SVC-based uncertainty}
        \label{fig:2c}
    \end{subfigure}
    \hfill
    \begin{subfigure}{0.48\textwidth}
            \centering
            \includegraphics[width=\linewidth]{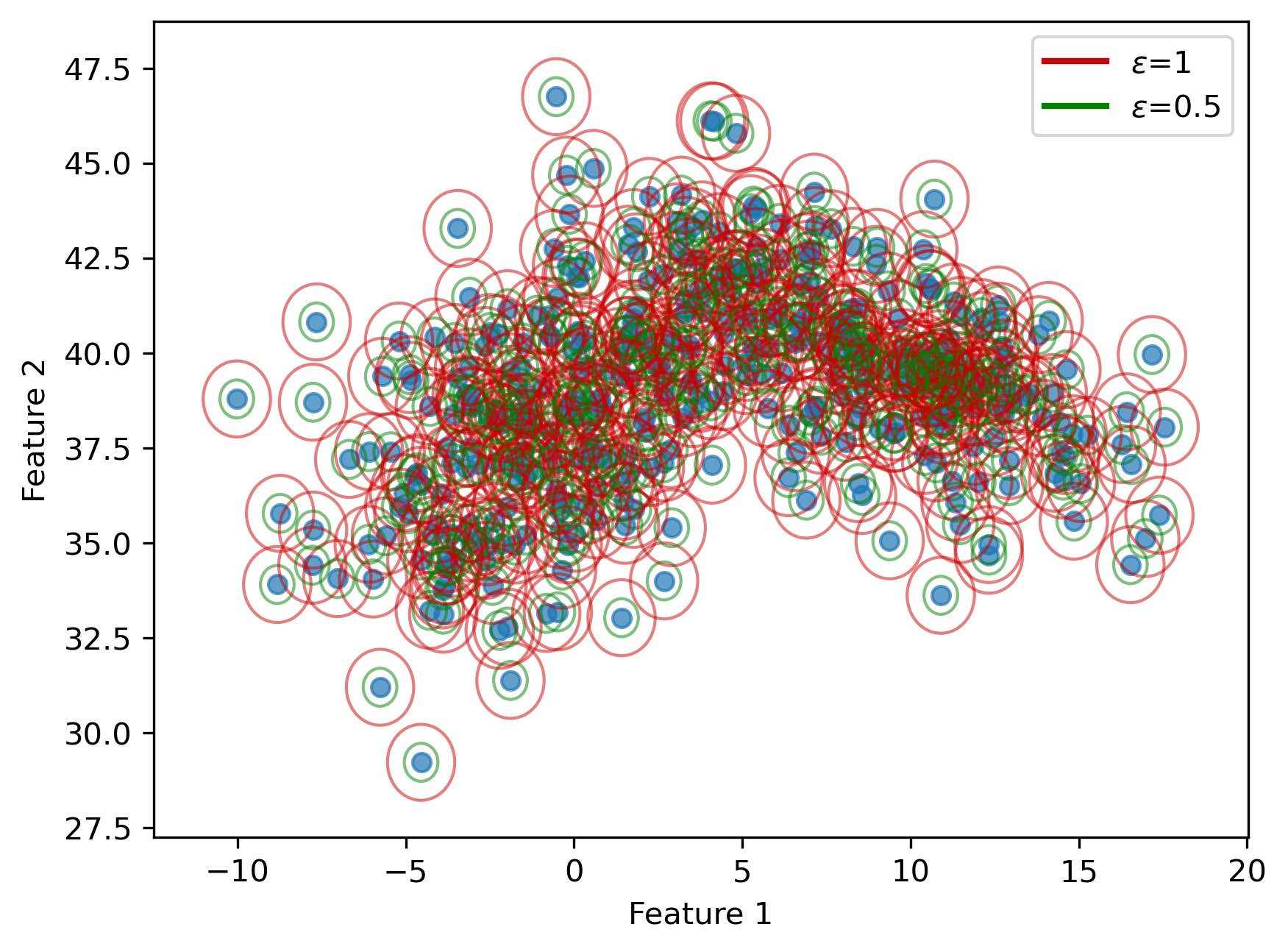}
            \caption{DRSP ambiguity}
            \label{fig:2d}
    \end{subfigure}
    \caption{Visual examples of uncertainty and ambiguity set geometries for four robust methods using a randomly generated two-dimensional dataset}
    \label{fig:2}
    \end{figure}

In this section, four robust route planning approaches are employed for sidewalk robot delivery, generating three different types of uncertainty sets and one ambiguity set. A visual example for each approach using a randomly generated dataset is shown in Figure~\ref{fig:2}. This dataset contains 400 two-dimensional samples generated from a mixture of Gaussian distributions. The two axes (Feature 1 and Feature 2) represent abstract cost features in a low-dimensional feature space and are used only for visualization. They do not correspond to specific sidewalk segments, travel times, or physical quantities, and therefore have no physical units. 

The parameters $\Gamma$, $\lambda$, and $v$ control the shape and size of the uncertainty sets by regulating the level of bounded deviation and model conservatism. In contrast, the parameter $\epsilon$ governs the size of the ambiguity set in the DRSP formulation. Instead of defining a single deterministic boundary in the feature space, each sample in the ambiguity set is associated with a Wasserstein ball, representing a family of plausible probability distributions. This figure is intended solely for conceptual illustration, highlighting the structural differences among the methods rather than conveying quantitative performance insights.

\subsubsection{Budgeted uncertainty}

The Budgeted uncertainty approach (\citet{Bertsimas2003}) for robust discrete optimization involves defining each entry \({u}_j \),  \(j \in [n]\) within the interval \([\underline{c}_j , \underline{c}_j+{d_j}]\). In the case of RSPP, \({u}_j \) represents the robust cost at certain segment \(j \), \(\underline{c}_j\) is the minimum observed cost of segment \(j\), and \({d_j}\) denotes the deviation between maximum cost and minimum cost. The degree of conservatism is controlled by the parameter \( \Gamma\) (Figure ~\ref{fig:2a}) which restricts at most  \( \Gamma \in \{1,2,...,n\}\) to take values in the interval, while the remaining entries take the minimum values. If \( \Gamma = 0\), the influence of the cost deviations is completely ignored. The budgeted uncertainty set is expressed as:
\begin{equation}\label{eq:3}
U^B=  \left\{ \mathbf{u} \middle| 
\begin{aligned}
    & \underline{c}_j \leq{u}_j \leq \underline{c}_j +{d_j}\lambda_j,\forall j \in [n]
\\  & \sum_{j \in [n]} \lambda_j \leq \Gamma, \lambda_j \in \{0,1\}
\end{aligned}   
\right\} 
\end{equation}
This problem can be formulated as:
\begin{equation}\label{eq:4}
\begin{aligned}
{\min} \quad & \mathbf{\underline{c}}\mathbf{x}+\underset{\{ S \mid S \subseteq [n], \ |S| \leq \Gamma \}}{\max} \sum_{(p,q) \in S} {d_{pq}}{x_{pq}}\\
\text{s.t.} \quad & 
\mathbf{x} \in X
\end{aligned}
\end{equation}

The constraints here refer to the shortest path problem. Finding such a robust solution is NP-hard, but this problem can be solved by solving at most \(n+1\) nominal shortest path problems \citep{Bertsimas2003}. This algorithm allows modelers to balance cost and robustness by adjusting the parameter \( \Gamma\). 

\subsubsection{Ellipsoidal uncertainty}

The Ellipsoidal uncertainty \citep{BENTAL1998,BENTAL1999} encompasses various reasonable types of ellipsoids and the intersections of finitely many ellipsoids. They make the corresponding robust convex program a tractable problem. An ellipsoidal uncertainty set can be in the form of
\begin{equation}\label{eq:5}
U^E = \left\{\mathbf{u}:(\mathbf{u} -\mathbf{\hat{c}})^T \mathbf{\Sigma}^{-1} (\mathbf{u} -\mathbf{\hat{c}}) \leq    \lambda \right\}
\end{equation}
where \(\lambda \geq 0\) is a parameter controlling the size of the ellipsoid (Figure ~\ref{fig:2b}). Here \(\mathbf{\hat{c}} \) and \(\mathbf{\Sigma}\) are generated from observed points considering the best fit of multivariate normal distribution \(N(\mathbf{\hat{c}},\mathbf{\Sigma})\):
\begin{equation}\label{eq:6}
\mathbf{\hat{c}} = \frac{1}{N}\sum_{i \in [N]}\mathbf{c}^i
\end{equation}
\begin{equation}\label{eq:7}
\mathbf{\Sigma} = \frac{1}{N}\sum_{i \in [N]} (\mathbf{c}^i-\mathbf{\hat{c}}) (\mathbf{c}^i-\mathbf{\hat{c}})^T
\end{equation}
In ellipsoidal uncertainty, they represent the center of the ellipsoid as well as the shape and the size of the ellipsoid respectively. The resulting robust shortest path problem under ellipsoidal uncertainty can be denoted as \citep{dokka2017}:
\begin{equation}\label{eq:8}
\begin{aligned}
{\min} \quad & \mathbf{\hat{c}}\mathbf{x}+z\\
\text{s.t.} \quad &  z^2  \geq \lambda (\mathbf{x}^t \Sigma \mathbf{x})\\
 \quad & 
\mathbf{x} \in X
\end{aligned}
\end{equation}
The term \(z\) here is an auxiliary variable to account for the worst-case impact of the uncertainty set.

\subsubsection{Kernel-based SVC uncertainty}
\label{sec:svc}

The SKL-based SVC uncertainty approach \citep{SHANG2017464} is a data-driven approach that combines an unsupervised machine learning algorithm, the SVC, with the Weighted Generalized Intersection Kernel (WGIK) method. It utilizes Support Vectors (SV) to define boundaries in feature space, grouping data points into clusters by finding the smallest sphere that encloses them. The kernel function WGIK is calculated in Eq. \eqref{eq:9}:
    \begin{equation}\label{eq:9}
    K(\mathbf{u},\mathbf{v})=\sum_{k=1}^n l_k -||\mathbf{Q}(\mathbf{u}-\mathbf{v})||_1
    \end{equation}
    where \(l_k\) are kernel scaling constraints, \(\mathbf{Q}\) is a weighting matrix generated from covariance matrix  \(\mathbf{Q}=\mathbf{\Sigma}^{-\frac{1}{2}}\), and \(\mathbf{u}\), \(\mathbf{v}\) represent two data points. This kernel function measures similarity between data points and is concave, ensuring a convex acceptance region for the SVC.

This SKL-based SVC approach not only manages correlated uncertainties, resulting in asymmetric uncertainty sets, but it also features adaptive complexity, embodying a nonparametric approach. The resulting convex polyhedral uncertainty set can ensure the robust counterpart problem of the same type as the deterministic problem. The explicit expression of this uncertainty set can be derived as a set of linear inequalities:
    \begin{equation}\label{eq:10}
    U^{SVC}=  \left\{ \mathbf{u} \middle| 
    \begin{aligned}
        & \exists \mathbf{v}_i, i \in SV  s.t.
    \\  & \sum_{i \in SV} \alpha_i \cdot \mathbf{1}^T \mathbf{v}_i \leq \theta
    \\  & -\mathbf{v}_i \leq \mathbf{Q}(\mathbf{u}-\mathbf{u}^i) \leq \mathbf{v}_i, \forall i \in SV
    \end{aligned}   
    \right\} 
    \end{equation}
    where \( \alpha_i\) are Lagrangian multipliers obtained from the SVC dual problem, \(\mathbf{v_i}\) are auxiliary variables, and \(\theta\) defines the SVC boundary and is obtained from the SVs. \( SV\) refers to a list of support vectors, the number of which is controlled by a predefined regularization parameter \(v \in (0,1]\). It is an upper bound on the fraction of outliers and a lower bound on the fraction of support vectors (Figure ~\ref{fig:2c}). It is introduced to control the conservatism degree of the uncertainty set.

Before applying the above SKL-based SVC uncertainty set into the robust shortest path problem, Eq.\eqref{eq:2} can be first simplified by moving the inner objective function into a constraint:
    \begin{equation}\label{eq:11}
    \begin{aligned}
    \underset{\mathbf{x}\in X, b}{\min} \quad & b \\
    \text{s.t.} \quad & \underset{\mathbf{u}\in U}{\max} \mathbf{u}^T \mathbf{x} \leq b  
    \end{aligned}
    \end{equation}
 where the auxiliary variable \(b\) represents an upper bound on the worst-case path cost over the uncertainty set and serves as the epigraph variable of the robust objective. 

Then the linear inequality constraints in Eq.\eqref{eq:10} can replace the uncertainty set  \(U\) in the left-hand side (LHS) of the constraint in Eq. \eqref{eq:11}. After introducing Lagrange multipliers \(\mathbf{\lambda_i}\), \(\mathbf{\mu_i}\), and \(\eta\), the final model for the robust shortest path problem with WGIK-SVC-based uncertainty set can be defined as follows:
\begin{equation}\label{eq:12}
\begin{aligned}
\underset{x,\mathbf{\lambda_i},\mathbf{\mu_i},\mathbf{\eta}}{\min} \quad & b 
\\ \text{s.t.} \quad &  
\sum_{i \in SV} (\mathbf{\mu_i} - \mathbf{\lambda_i})^T \mathbf{Q}\mathbf{u}^i + \eta\theta \leq b 
\\ \quad & \sum_{i \in SV} \mathbf{Q}(\mathbf{\lambda_i} - \mathbf{\mu_i}) + \mathbf{x}  =  \mathbf{0} 
\\ \quad &   \mathbf{\lambda_i} + \mathbf{\mu_i} = \eta \cdot \alpha_i \cdot \mathbf{1},  \forall i \in SV 
\\ \quad & \mathbf{\lambda_i},\mathbf{\mu_i}\in \mathbb{R}_+^n, \eta \geq 0, \mathbf{x} \in X
\end{aligned}
\end{equation}
The detailed introduction of SKL-based SVC and derivation of Eq.\eqref{eq:12} is provided in Appendix~\ref{app:svc-derivation}. 

This WGIK-based SVC approach still achieves high computational efficiency by formulating the robust counterpart problem as the deterministic problem even if it generates \(2*|SV|*|A|+1\) additional variables. However, its computational tractability is lost when applied to the RSPP. The RSPP with kernel-based SVC uncertainty is Mixed-Integer Programming (MIP) with both continuous variables (Lagrange multipliers) and binary variables (path choices), and usually has many data points (large \(|SV|\)) and high dimensions (large \(|A|\)). This makes this problem computationally intensive and hard to find a solution efficiently even under a small road network. In addition, the SVC is impacted by the curse of dimensionality, suggesting that the accuracy of the resultant data-driven uncertainty sets may decline when dealing with high-dimensional uncertainties \citep{NING2018190}. This is particularly evident in RSPP because of high dimensions. 

To tackle the aforementioned challenges, the Two-Stage Clustering with Dimensional Separation (TSC-DS) algorithm proposed by \citet{ROYTVANDGHIASVAND2024} is employed. Unlike other methods to overcome the curse of dimensionality, such as Principal Component Analysis (PCA) which is used for highly correlated datasets as described in \citep{SHANG201819}, the TSC-DS algorithm is specifically developed for column-wise uncertainty. This algorithm not only significantly reduces the number of SVs, but also decreases the number of dimensions. At the first stage, it eliminates additional SVs that do not contribute to the uncertainty set construction using the SKL-based SVC approach. The remaining dataset is then divided into \(F\) low-dimensional subsets. At the second stage, the SVC uncertainty set \(U^{SVC}_f, f \in [F]\) for each subset is constructed. Each uncertainty set \(U^{SVC}_f\) can be written as:
\begin{equation}\label{eq:13}
U^{SVC}_f=  \left\{ \mathbf{u}_f \middle| 
\begin{aligned}
    & \exists \mathbf{v}_{i_f}, {i_f} \in SV_f      s.t.
\\  & \sum_{{i_f} \in SV_f} \alpha_{i_f} \cdot \mathbf{1}^T \mathbf{v}_{i_f} \leq \theta_f
\\  & -\mathbf{v}_{i_f} \leq \mathbf{Q}_f(\mathbf{u}_f-\mathbf{u}^{i_f}) \leq \mathbf{v}_{i_f}, \forall {i_f} \in SV_f
\end{aligned}   
\right\} 
\end{equation}
Then the final model Eq.\eqref{eq:14} can be reformulated as:
\begin{equation}\label{eq:14}
\begin{aligned}
\underset{x,\mathbf{\lambda}_{i_f},\mathbf{\mu}_{i_f},\mathbf{\eta}_f}{\min} \quad & b 
\\ \text{s.t.} \quad &  
\sum_{f \in [F]} \left( \sum_{{i_f} \in SV_f} (\mathbf{\mu}_{i_f} - \mathbf{\lambda}_{i_f})^T \mathbf{Q}_f\mathbf{u}^{i_f} + \eta_f\theta_f \right) \leq b 
\\ \quad & \sum_{i_f \in SV_f} \mathbf{Q}_f(\mathbf{\lambda}_{i_f} - \mathbf{\mu}_{i_f}) + \mathbf{x}_f  =  \mathbf{0} , \forall f \in [F] 
\\ \quad &   \mathbf{\lambda}_{i_f} + \mathbf{\mu}_{i_f} = \eta_f \cdot \alpha_{i_f} \cdot \mathbf{1},  \forall {i_f} \in SV_f , \forall f \in [F] 
\\ \quad & \mathbf{\lambda}_{i_f},\mathbf{\mu}_{i_f}\in \mathbb{R}_+^n, \eta_f \geq 0, \mathbf{x} \in X 
\end{aligned}
\end{equation}

It is worth mentioning that the dimensional separation technique employed in the TSC-DS algorithm significantly enhances the applicability of the MKL-based SVC approach. This approach is particularly suitable for managing uncertainty in lower-dimensional spaces, specifically when the number of dimensions is fewer than eight \citep{HAN2021}. The advantage of the MKL method lies in its ability to flexibly select and combine multiple kernel functions, optimizing the kernel weights \(\pi_{m}\) automatically to best fit the data distribution. In contrast, the SKL method relies on a single pre-specified kernel, which may not sufficiently capture complex structures in high-dimensional or non-linearly separable data. 

\citet{HAN2021} proposed a concave kernel structure for MKL as an extension of the WGIK function, enabling a more refined construction of uncertainty sets in a robust optimization (RO) framework. The proposed concave kernel function is defined as follows:
\begin{equation}\label{eq:15}
    K_m(\mathbf{u},\mathbf{v})=1-\left|\frac{\mathbf{q}_m^{\top} (\mathbf{u} - \mathbf{v})}{c_m \cdot \kappa}\right|
    \end{equation}
where \(\mathbf{q}_m\)  represents the projection direction, \(c_m\) is a normalization factor, \(\kappa\) controls the scaling of the kernel function, and \(m \in [M]\), with \(M\) representing the number of kernels. 

A critical factor in MKL is the regularization parameter \(\mu\), which directly controls how many kernels are actively used in constructing the uncertainty set. A lower \(\mu\) value encourages sparsity, forcing the optimization process to select only a few relevant kernels while setting others to zero. A higher \(\mu\) value distributes the weight across multiple kernels, increasing complexity.

Building on this foundation, the RO-compatible MKL-based SVC uncertainty set in the second stage of the TSC-DS algorithm can be yielded as:
\begin{equation}\label{eq:16}
U^{MKL-SVC}_f=  \left\{ \mathbf{u}_f \middle| 
\sum_{i_f \in SV_f} \alpha_{i_f}^{\star} \sum_{m_f \in SK_f} \pi_{m_f}^{\star} K_{m_f}(\mathbf{u}_f, \mathbf{u}_{i_f}) \geq \rho_f^{\star}
\right\} 
\end{equation}
where \(SK_f\) is the set of selected supported kernels for subset \(f\), \(\alpha_{i_f}^{\star}\) are the optimized dual variables associated with the SVs, \(\pi_{m_f}^{\star}\) are the optimized kernel weight coefficients, which determine the relative importance of different basis kernels, and \(\rho_f^{\star}\) is a threshold parameter defining the boundary of the uncertainty set.

Consequently, the RO model for RSPP using MKL-based SVC uncertainty can finally be reformulated as:

\begin{equation}\label{eq:17}
\begin{aligned}
\underset{x,\mathbf{\lambda}_{i_f},\mathbf{\mu}_{i_f},\mathbf{\eta}_f}{\min} \quad & b 
\\ \text{s.t.} \quad &  
\sum_{f \in [F]} \left( \sum_{{i_f} \in SV_f}\sum_{m_f \in SK_f} (\mu_{i_f m_f } - \lambda_{i_f m_f})\frac{\mathbf{q}_{m_f}^{\top} (o_{i_f})}{c_{m_f} \cdot \kappa_f}   + \eta_f(1-\rho_f^{\star})\right) \leq b 
\\ \quad & \sum_{i_f \in SV_f} \sum_{m_f \in SK_f}(\lambda_{i_f m_f} -\mu_{i_f m_f})\frac{\mathbf{q}_{m_f}}{c_{m_f} \cdot \kappa_f} + \mathbf{x}_f  =  \mathbf{0} , \forall f \in [F] 
\\ \quad &   \lambda_{i_f m_f} + \mu_{i_f m_f} = \eta_f \cdot \alpha_{i_f}^{\star} \cdot \pi_{m_f}^{\star},  \forall {i_f} \in SV_f , m_f \in SK_f ,  \forall f \in [F] 
\\ \quad & \lambda_{i_fm_f},\mu_{i_fm_f}\geq 0,\forall {i_f} \in SV_f , m_f \in SK_f ,  \forall f \in [F] 
\\\quad & \eta_f \geq 0 ,\forall f \in [F] , \mathbf{x} \in X
\end{aligned}
\end{equation}

\subsubsection{DRSP ambiguity}

The other data-driven approach adopted is the DRSP method using Wasserstein ambiguity sets proposed by \citet{wang2019,wang2020}. Unlike the aforementioned models, this method does not rely on uncertainty sets capturing all possible variations. Instead, it defines an ambiguity set (i.e., a Wasserstein ball) containing all probability distributions within a specified Wasserstein distance \(\epsilon_N\) from the empirical distribution. This ambiguity set is derived from finite samples and helps address the uncertainty in travel times. This formulation offers a flexible way to incorporate data-driven uncertainty and control conservatism through the radius \(\epsilon_N\), which reflects the confidence in the empirical data (Figure ~\ref{fig:2d}). The optimal path is determined by minimizing the worst-case \(\alpha\)-reliable mean-excess travel time (METT), which represents the expected travel time exceeding the \(\alpha\)-quantile and accounts for the risk of extreme delays, evaluated over all distributions in the Wasserstein ambiguity set. The formulation of the \(\alpha\)-METT for a given path \(\mathbf{x}\) under the distribution \(F\) is given below:

\begin{equation}\label{eq:18}
\mathrm{METT}_\alpha(\mathbf{x})
=\min_{t\in\mathbb{R}}
\Bigl\{\,t \;+\;\frac{1}{\alpha}\,\mathbb{E}_F\bigl[\max\{\boldsymbol{\xi}^T \mathbf{x} - t,\,0\}\bigr]\Bigr\}
\end{equation}
where \(\boldsymbol{\xi} \in\mathbb{R}^n \) is the vector of segment travel times under distribution \(F\), \(t\) is the quantile threshold, and \(\max\{\boldsymbol{\xi}^T \mathbf{x} - t,\,0\}\) measures the excess travel time above threshold \(t\). 

Furthermore, this method incorporates support intervals to ensure that perturbed distributions remain realistic. Let \([\mathbf{a},\mathbf{b}]\) denote the empirical support interval for segments, constructed from the free flow and maximum observed travel times. The Wasserstein DRSP model can be formulated as:  

\begin{equation}\label{eq:19}
\begin{aligned}
\min_{t, \mathbf{s}, \boldsymbol{\lambda}, \boldsymbol{\gamma}_i, \boldsymbol{\eta}_i} \quad & 
t + \frac{1}{\alpha} \left\{ \frac{1}{N} \sum_{i=1}^N s_i + \lambda \epsilon_N \right\} \\
\text{s.t.} \quad 
& \left( \mathbf{x} + \boldsymbol{\gamma}_i - \boldsymbol{\eta}_i \right)^T \hat{\boldsymbol{\xi}}^i 
- \boldsymbol{\gamma}_i^T \mathbf{a} + \boldsymbol{\eta}_i^T \mathbf{b} - t \leq s_i \\
& \| \boldsymbol{\gamma}_i + \mathbf{x} - \boldsymbol{\eta}_i \|_q \leq \lambda \\
& \boldsymbol{\eta}_i \geq 0,\quad \boldsymbol{\gamma}_i \geq 0,\quad s_i \geq 0, 
\quad \forall i \in [N] \\
& \mathbf{x} \in X
\end{aligned}
\end{equation}
where \(\hat{\boldsymbol{\xi}}^i\) is the \(i\)-th sample vector, \(N\) denotes the total number of empirical travel-time samples used to construct the empirical distribution and the Wasserstein ambiguity set, \(\boldsymbol{\eta}_i\) and \(\boldsymbol{\gamma}_i\) are Lagrange multipliers for the support set, \(s_i\) is a slack variable, and \(\|\cdot\|_q\) is the dual norm corresponding to the chosen \(l_p\)-norm for the Wasserstein distance. The reformulation of the worst-case METT problem into a tractable mixed 0--1 convex program allows for efficient optimization even under high-dimensional and sample-based uncertainty.

The two key parameters, \(\epsilon_N\) and \(\alpha\), play crucial roles in controlling the model’s robustness and risk sensitivity. The radius \(\epsilon_N\) governs the size of the Wasserstein ball: increasing \(\epsilon_N\) leads to a more conservative model by expanding the ambiguity set and allowing for greater deviations from the empirical distribution. Conversely, a smaller \(\epsilon_N\) tightens the ambiguity set and reduces robustness, relying more heavily on the quality of the data. The parameter \(\alpha\), which appears in the definition of the METT, represents the tail probability level. A smaller \(\alpha\) results in a more risk-averse model that penalizes large deviations and extreme outcomes more heavily, potentially increasing the total cost. Larger \(\alpha\) values yield a model that is less sensitive to outliers, favoring average-case performance. Therefore, proper tuning of \(\epsilon_N\) and \(\alpha\) is essential for balancing robustness, reliability, and computational efficiency.

\subsection{Sidewalk robot simulation}
\label{sec:sidewalkrobot simulation}

To simulate realistic sidewalk navigation for sidewalk delivery robots, we integrate an agent-based pedestrian simulator, PTV VISWALK \citep{ptv2023}, that accounts for interactions among pedestrians, robots, and static urban obstacles. VISWALK is based on the Social Force Model (SFM), which reproduces pedestrians movements and interactions at the microscopic level. 

The SFM, one of the most prominent models for pedestrian dynamics, was originally proposed by \citet{helbing1995social}. It describes a pedestrian's motion in the form of an acceleration or deceleration, resulting from a number of different forces acting on the pedestrian such as social, psychological, and physical forces. Given a certain pedestrian, the total force \(\vec{F}\) can be expressed as the sum of four terms \citep{ptv2023}:
\begin{equation}\label{eq:20}
\vec{F}= \vec{F}_{driving} + \vec{F}_{social} + \vec{F}_{wall} + \vec{F}_{noise}
\end{equation}
where \(\vec{F}_{driving}\) represents the driving force that the pedestrian wants to reach a certain destination as comfortable as possible, \(\vec{F}_{social}\) refers to the interaction forces between the pedestrian and other individuals, \(\vec{F}_{wall}\) represents the interaction forces between the pedestrian and obstacles, \(\vec{F}_{noise}\) is a random force that added to the total social forces if a pedestrian remains below his desired speed for a certain time, to avoid simulated pedestrians getting stuck for periods of time.

\(\vec{F}_{driving}\) is described by an acceleration term of the form:
\begin{equation}\label{eq:21}
 \vec{F}_{driving} = \frac{\vec{v}_{0}-\vec{v}}{\tau}
\end{equation}
In this formulation,  \(\vec{v}_{0}\) is the desired speed of the individual, while \(\vec{v}\) is the instantaneous speed. The term \(\tau\) refers to the relaxation time required for the individual to approach the desired speed. A larger \(\tau\) implies that it takes the pedestrian more time to adjust its velocity. 

The original \(\vec{F}_{social}\) was extended by \citet{johansson2008specification} who introduced an angle-related friction parameter to express the anisotropy in the model. This modification has been implemented into VISWALK's SFM by splitting the original \(\vec{F}_{social}\) as the sum of isotropic \(\vec{F}_{soc\_iso}\) and mean social forces \(\vec{F}_{soc\_mean}\) to consider the the effect of vision field and anisotropy. The isotropic social force that the pedestrian \(\alpha\) experiences from another pedestrian \(\beta\) is defined as follows:
\begin{equation}\label{eq:22}
 \vec{F}_{\alpha\beta}^{soc\_iso} = A_{soc\_iso}w(\lambda)e^{-d_{\alpha\beta}/B_{soc\_iso}}\vec{n}_{\alpha\beta}
\end{equation}
where \(A_{soc\_iso}\)  and \(B_{soc\_iso}\) are non-measurable parameters controlling the two forces among pedestrians, and \(d\) is the distance between the pedestrian \(\alpha\) and the influencing pedestrian \(\beta\). The term \(\vec{n}\) refers to the unit vector pointing from the influencing pedestrian \(\beta\) to the influenced pedestrian \(\alpha\).  The \(\lambda\) is an anisotropy factor that makes the forces from behind not influence the pedestrian as much as forces in the front. It ranges from 0 to 1 and grows with the strength of interactions from behind. By the function \(w(\lambda)\) below, the force is dependent of direction:
\begin{equation}\label{eq:23}
 w(\lambda) = \left( \lambda + (1- \lambda) \frac{1+\cos(\varphi)}{2} \right)
\end{equation}
The \(\varphi\) denotes the angle between the current direction of the pedestrian \(\alpha\) and the relative position of the influencing pedestrian \(\beta\). The overall isotropic social force for pedestrian \(\alpha\) is the sum of \(\vec{F}_{\alpha\beta}^{soc\_iso}\) for all nearby pedestrians \(\beta\) within the range identified by the parameter \textit{react\_to\_n}, which is used to control the maximum number of persons considered in the closest surrounding.

The formula of mean social forces is similar to that of isotropic social forces:
\begin{equation}\label{eq:24}
 \vec{F}_{\alpha\beta}^{soc\_mean} = A_{soc\_mean}e^{-d_{\alpha\beta}/B_{soc\_mean}}\vec{n}_{\alpha\beta}
\end{equation}
where \( A_{soc\_mean}\) and \( B_{soc\_mean}\) denote the strength and range of the social force respectively. However, all forces that act in the back (\(180^\circ\)) of the pedestrians are ignored. Moreover, a parameter \(VD\) is introduced to generalize the distance \(d_{\alpha\beta}\) in order to take into account the step width of the influencing pedestrian. If \(VD\) = 0, it will not consider the relative velocities of pedestrians and is equal to the body surface distance between two pedestrians.

\(\vec{F}_{wall}\) represents the interaction forces between the pedestrian and obstacles, and is calculated in the same way as \(\vec{F}_{social}\). \(\vec{F}_{noise}\) refers to the fluctuation term (Noise) in the original SFM, which is expressed through an acceleration parameter \(m/s^2\).

Although the force terms above define how agents respond to goals, other agents, and obstacles, the SFM was originally developed to reproduce pedestrian motion and does not explicitly encode vehicle-level kinematic constraints. Sidewalk delivery robots may differ substantially from pedestrians in their maneuverability, particularly in terms of turning behavior and lateral motion when avoiding obstacles. In VISWALK, such differences are not enforced through explicit constraints (e.g., minimum turning radius or steering limits), but instead emerge implicitly from the chosen SFM parameters. As a result, different parameterizations that are all plausible for pedestrians may lead to robot trajectories with very different implied kinematic performance, which motivates the need to explicitly examine the kinematic implications of the selected SFM parameters when modeling sidewalk robots. 

Given the lack of established SFM calibration for sidewalk delivery robots in VISWALK, we rely on prior studies that apply or adapt the SFM to robot navigation in pedestrian environments to guide parameter selection. \citet{Ferrer2013robotcompanion, Ferrer2013robotnavigation} introduce a social-aware navigation approach by applying the SFM to the robot itself to govern its movement. Meanwhile, research by \citet{truong2017toward} develop an extended SFM for mobile robot navigation in dynamic and crowded environments. Following these studies, we focus on four parameters that most strongly influence robot behaviors:  \(\tau\), \(A_{soc\_iso}\), \(B_{soc\_iso}\), and \(\lambda\). The tested parameter ranges are selected to be consistent with values commonly used in previous studies: \(\tau \in \{0.4,0.8,1.2\}\), \(A_{soc\_iso} \in \{1,2,3\}\), \(B_{soc\_iso} \in \{0.2,0.35,0.5\}\), \(\lambda \in \{0.2,0.45,0.7\}\). These values allow us to explore differences in responsiveness and avoidance behavior without introducing extreme or unstable motion patterns. Additionally, the desired speed of sidewalk robots is fixed at 5 km/h with no deviation, and their size is set to the default size of an adult male in VISWALK, which is 0.54 m wide.

To assess how these 81 parameter combinations reflect robot maneuverability, we conduct a dedicated kinematic sensitivity analysis in a controlled micro-scenario, consisting of a straight sidewalk segment with two static obstacles placed to require lateral avoidance. In this setup, a single robot traverses the segment repeatedly under identical pedestrian environmental conditions, so that differences in motion arise only from the SFM parameters rather than from varying pedestrian interactions. For each parameter combination, we run 10 replications with different random seeds and extract the resulting trajectories from VISWALK. From these trajectories, we compute kinematic performance indicators that reflect turning behavior and motion smoothness, including implied turning radius, yaw-rate and lateral-acceleration proxies derived from trajectory curvature, as well as total travel time and distance. The results of all 81 parameter combinations can be found in Appendix~\ref{app:sfm_full_results}. For reference, we also report the VISWALK default pedestrian parameterization as a baseline (shown in bold in the first row of the appendix table). By reporting the kinematic performance ranges observed across all parameter sets, we explicitly characterize how much robot maneuverability can vary under plausible SFM configurations, instead of assuming that pedestrian-calibrated parameters directly represent robot motion. 

Based on this kinematic sensitivity analysis, we select one parameter combination to represent a default robot navigation profile for the main experiments. This parameter set lies in the central region of the observed kinematic ranges and does not exhibit extreme turning aggressiveness or unstable motion. In the micro-scenario, this profile yields a 5th-percentile implied turning radius of 1.17 m, a 95th-percentile yaw rate of 0.73 rad/s, and a 95th-percentile lateral-acceleration proxy of 0.71 m/s\textsuperscript{2}, which are consistent with moderate and smooth maneuvering behavior under the speed limit of 5 km/h. This parameter set is therefore adopted as the baseline robot configuration reported in Table~\ref{tab:1}. 

Considering that the robot design specifications and operational settings may vary across manufacturers, two alternative parameter sets are retained to represent more conservative and more aggressive maneuvering behavior, respectively. These alternatives are evaluated in Section~\ref{sec: sensitivity_analysis} to examine how variations in robot behavior affect the performance of robust routing strategies, ensuring that the main conclusions are not driven by a single, unexamined motion assumption. 

\begin{table*}[!h]
        \centering
        \caption{Social force model parameters for pedestrians and robots}
        \footnotesize
        \begin{tabular}{|c|c|c|c|c|c|c|c|c|c|}
        \hline  
        \textbf{Parameter}&$\boldsymbol{\tau}$&\textbf{ReactToN} & 
        $\boldsymbol{A}_{\text{soc\_iso}}$ & 
        $\boldsymbol{B}_{\text{soc\_iso}}$ & 
        $\boldsymbol{\lambda}$ &
        $\boldsymbol{A}_{\text{soc\_mean}}$ & 
        $\boldsymbol{B}_{\text{soc\_mean}}$ & 
        \textbf{VD} &
        \textbf{Noise}\\ \hline 
        
        \textbf{Pedestrian}& 0.4&  8& 2.72& 0.2& 0.176& 0.4& 2.8& 3&1.2\\ \hline 
        \textbf{Robot}& 0.8&  8& 2.0& 0.35& 0.45& 0.4& 2.8& 3&1.2\\ \hline 
        \end{tabular}
    \label{tab:1}
    \end{table*}

\subsection {Implications for multi-stop tours}
\label{sec:implications_multi_stop}

In multi-stop settings, the sequence of customers is typically optimized by using arc costs between pairs of stops, for example in TSP or VRP formulations. While this paper does not explicitly address a tour-level optimization problem, the generation of uncertainty-aware costs inevitably affects the solutions. Uncertainty-aware path costs that are embedded as arc costs in tour planning problems allow for more robust tours.
To clarify this effect, we provide an analytical illustration showing how travel-time uncertainty propagates along a fixed delivery sequence, and how replacing nominal shortest-path legs with robust shortest-path legs affects the arrival delays at ''downstream'' customers.

Consider a fixed tour visiting $n$ customers in a predetermined order. Let leg $r$ denote the movement between any two consecutive stops. The arrival time at customer $n$ is given by the cumulative sum of leg travel times:

\begin{equation}
A_n = \sum_{r=1}^{n} T_r,
\qquad
A_n' = \sum_{r=1}^{n} T_r',
\end{equation}

where $T_r$ and $T_r'$ denote, respectively, the travel time on leg $r$ under nominal shortest path (SP) and robust shortest path (RSP). Each leg travel time can be decomposed into a free-flow travel time (invariant and depending only the robot speed) and a stochastic delay component:
\begin{equation}
T_r = t_r + d_r,
\qquad
T_r' = t_r' + d_r',
\end{equation}

Using the nominal free-flow baseline $\sum_{r=1}^{n} t_r$, the arrival delay at customer $n$ is defined as:
\begin{equation}
\Delta A_n := A_n - \sum_{r=1}^{n} t_r,
\qquad
\Delta A_n' := A_n' - \sum_{r=1}^{n} t_r.
\end{equation}

Substituting the above formulations yields: 
\begin{align}
\Delta A_n &= \sum_{r=1}^{n} d_r, \\
\Delta A_n' &= \sum_{r=1}^{n} (t_r' - t_r) + \sum_{r=1}^{n} d_r'.
\end{align}

Taking into consideration expectations the arrival delays become:
\begin{equation}
\mathbb{E}[\Delta A_n] = \sum_{r=1}^{n} \mathbb{E}[d_r],
\qquad
\mathbb{E}[\Delta A_n'] = \sum_{r=1}^{n} (t_r' - t_r) + \sum_{r=1}^{n} \mathbb{E}[d_r'].
\end{equation}

Therefore, the expected reduction in arrival delay at customer $n$ obtained by replacing SP legs with RSP legs result into: 
\begin{equation}
\label{eq:delay_reduction_n}
\mathbb{E}[\Delta A_n - \Delta A_n']
= \sum_{r=1}^{n} (\delta_r - \epsilon_r).
\end{equation}

where $delta_r$ and $epsilon_r$ represent the per-leg difference in expected delays and difference in travel time, respectively:

\begin{equation}
\delta_r := \mathbb{E}[d_r] - \mathbb{E}[d_r'],
\qquad
\epsilon_r := t_r' - t_r.
\end{equation}

The above formulation highlights the fundamental trade-off between SP and RSP. Robust routing yields a positive expected improvement at a customer $n$ whenever the cumulative difference in expected delays is higher than the cumulative difference in travel time. Importantly, the effect is cumulative along the tour: even modest positive net effects on individual legs can aggregate into substantial downstream improvements, because delays accrue additively across consecutive stops. Under approximately homogeneous conditions where $\delta_r-\epsilon_r \approx \Delta>0$ across legs, the expected improvement grows approximately proportionally with the number of visited customers, $\mathbb{E}[\Delta A_n-\Delta A_n'] \approx n\Delta$. In more heterogeneous environments, the accumulation is driven by where high-variability legs occur.

\section{Results}
\label{Sec:Results}

This section focuses on the analysis of robust navigation for sidewalk robot delivery. We first evaluate the methods on a controlled synthetic network in Section ~\ref{sec:synthetic_setup}, which allows pedestrian demand and obstacle intensity to be varied systematically.  We then applies the methods to a detailed pedestrian network in central Stockholm with realistic spatial structure and demand variability in Section~\ref{sec:scenario setup}. Section~\ref{sec:performance analysis} systematically evaluates the performance of robust methods, offering insights into the trade-off between robustness and efficiency. Section~\ref{sec: sensitivity_analysis} provides a comprehensive analysis of how design-related factors and environmental factors affect the efficiency of both traditional routing and robust routing solutions. Finally, Section ~\ref{sec:operational_insights} summarizes the main findings from the Stockholm experiments and discusses practical implications for robust planning in sidewalk robot operations.

\subsection{Synthetic network}
\label{sec:synthetic_setup}

\begin{figure}[t]
    \centering
        \includegraphics[width=0.7\linewidth]{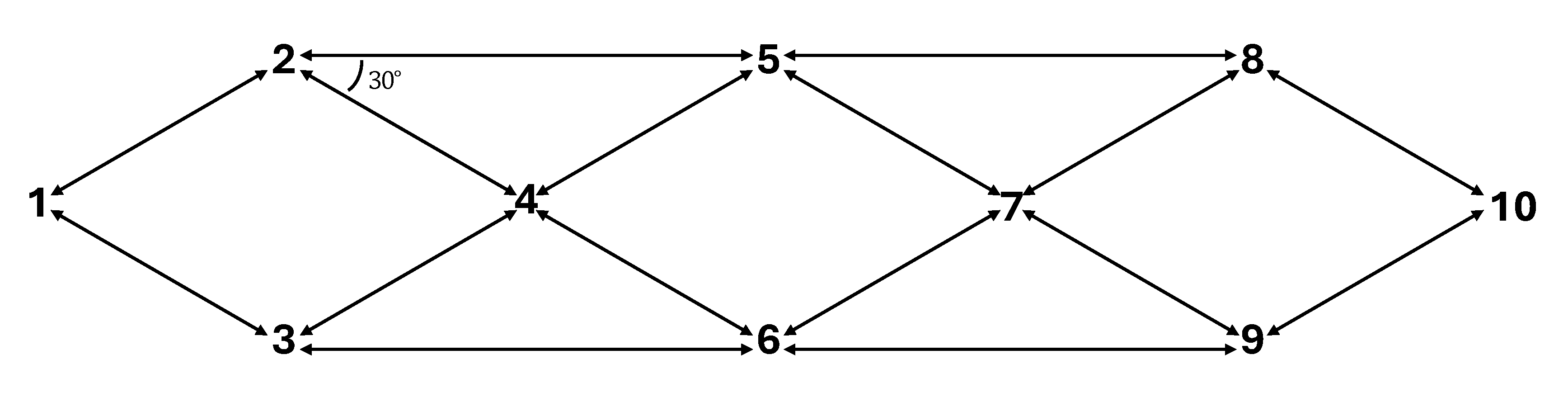}
    \caption{Synthetic network }
    \label{fig:3}
\end{figure}

We construct a controlled synthetic sidewalk network and generate a set of randomized simulation instances, to investigate the performance of robust navigation. The synthetic network consists of 10 nodes and 16 bidirectional sidewalk links, including 12 ``short links" of length 60~m and 4 ``long links" of length 104~m, as shown in Figure~\ref{fig:3}. All links have a uniform width of 3~m. The robot travels from a fixed origin at node~1 to a fixed destination at node~10.

The synthetic experiment is designed to preserve the key characteristics that drive uncertainty in sidewalk robot navigation, namely pedestrian flows and local obstacles. By construction, the conventional shortest path based solely on free-flow travel time selects one of the two boundary corridors containing long links, namely $1\!-\!2\!-\!5\!-\!8\!-\!10$ or $1\!-\!3\!-\!6\!-\!9\!-\!10$. However, these routes are made unreliable due to high pedestrian density and obstacle-induced bottlenecks, making this network a suitable benchmark for evaluating robust routing methods.

Three pedestrian flow levels are defined on the long links, based on the average flow criteria for pedestrian walkways in the Highway Capacity Manual \citep{HCM2010}, corresponding approximately to level‐of‐service (LOS) A-C:  (i) low demand: 1500~ped/h, (ii) medium demand: 3000~ped/h, and (iii) high demand: 4500~ped/h. To preserve a realistic contrast between boundary corridors and other links, the pedestrian flow on each short link is set to one third of the corresponding long-link flow level.

Three obstacle levels are also defined: low (3 obstacles), medium (6 obstacles), and high (9 obstacles). The obstacles are squares with side lengths randomly varying between 1 and 1.4 meters. Each obstacle is placed near the sidewalk edge at a random position along a randomly selected link, with at most one obstacle per link. This constraint avoids unrealistic full blockages while still creating meaningful bottlenecks and interaction effects.

The synthetic network is implemented in VISWALK using the robot behavioral model with parameters described in Section~\ref{sec:sidewalkrobot simulation}. The combination of pedestrian flow levels and obstacle levels result into 9 categories. For each category, we generate 20 replications. To emulate day-to-day stochastic fluctuations, the pedestrian input volumes in each replication are perturbed independently within $\pm 20\%$ of their category baseline value (while keeping the long- vs.\ short-link structure unchanged). This results in 180 simulated instances for the synthetic network. In addition, we compute the free-flow segment travel times as the ratio of segment length to the robot's desired speed and use the resulting conventional shortest path as a benchmark. All simulation instances and input data are publicly available via a GitHub repository\footnote{\url{https://github.com/xingtong1997/synthetic-sidewalk-robot-dataset.git}}.

The robust methods considered in this study include the Budgeted, Ellipsoidal, SKL- and MKL-based SVC, and DRSP approaches. Since the SVC is solved using the TSD-DS algorithm with random feature grouping, while sidewalk travel times are not entirely uncorrelated, we extended the TSC-DS algorithm by incorporating hierarchical clustering to form more meaningful groups. Specifically, the correlation matrix between segments is used as a distance matrix to perform hierarchical clustering. Segment pairs are then assigned based on these cluster labels, ensuring that each group contains two segments. This approach accounts for potential inter-segment correlations within each group and is expected to improve the performance of SVC compared to purely random grouping. Other clustering methods, such as K-means with variable group sizes, were tested but performed worse than random pairing and are therefore omitted. Both the extended SVC with hierarchical grouping and the original SVC with random grouping are included in the analysis.

Each robust method is evaluated across a range of parameter settings reflecting their theoretical definitions and practical relevance.  For the MKL-based SVC and DRSP methods, parameters are determined via preliminary experiments. For the MKL-based SVC method, we adopt \(M=16\) basis kernels and set \(\mu=0.2\). For the DRSP approach, we fix \(\alpha=0.2\), and vary \(\epsilon_N\). The full parameter ranges are:
\begin{itemize}
    \item Budgeted uncertainty: \( \Gamma \in \{1,2,...,10\}\);
    \item Ellipsoidal uncertainty: \( \lambda \in \{1,2,...,10\}\);
    \item SKL-based and MKL-based SVC uncertainty: \( v \in \{0.02,0.04,...,0.2\}\)
    \item Wasserstein-based DRSP:  \( \epsilon_N\in \{0.01,0.05,0.1,0.2,...,0.8\}\)
\end{itemize}

A five-fold validation is performed to ensure a rigorous performance evaluation. Instances are split into five folds using stratification over the 9 categories, ensuring each fold contains an equal number of replications from every category. In each fold, 80\% of instances are used as training data to construct uncertainty (or ambiguity) sets, and the remaining 20\% are used for validation.

To achieve a comprehensive evaluation of all methods, we adopt two performance criteria proposed by \citet{chassein2018} and define them as: 
\begin{itemize}
    \item the average travel time across all validation instances, 
    \item the average worst-case travel time across validation instances. 
\end{itemize}

For each validation instance, the realized travel time \(T\) of each method is computed and normalized by the free flow travel time of the benchmark path, \(T_{ff}\). The normalized delay $\tilde{D}$ corresponds to: \(\tilde{D}=(T-T_{ff})/T_{ff}\). Accordingly, the two performance criteria are referred to as average delay and worst-case delay. The first measures the typical operational efficiency under typical conditions, while the latter reflects protection against adverse realizations and can therefore be interpreted as robustness.

\begin{figure}[t]
    \centering
        \includegraphics[width=0.7\linewidth]{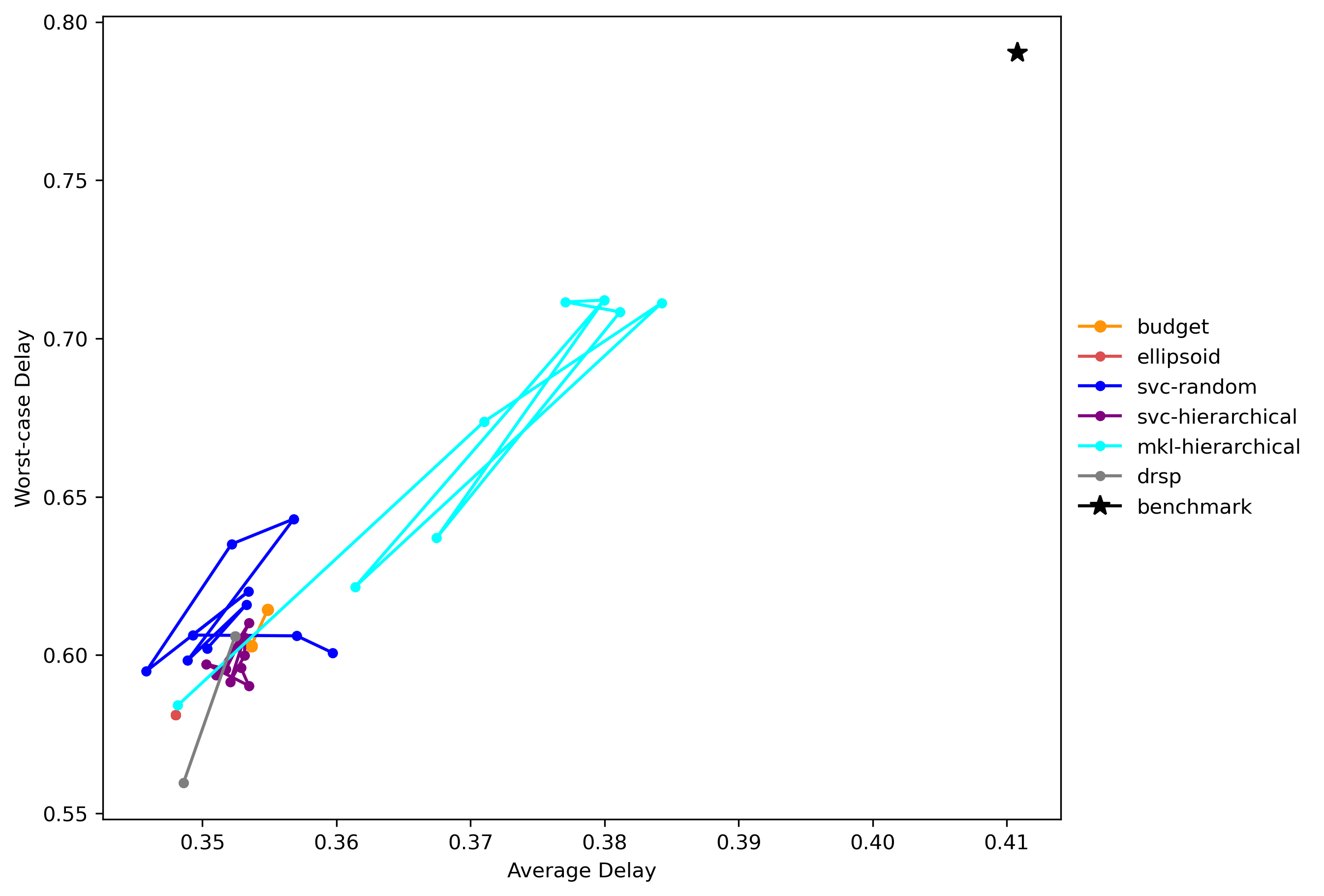}
    \caption{Average vs Worst case in Synthetic network}
    \label{fig:4}
\end{figure}

The performance of various methods under different parameter settings is illustrated in Figure~\ref{fig:4}. All robust methods outperform the benchmark in both average and worst-case delay, indicating improved efficiency and robustness. The DRSP method achieves the lowest worst-case delay (0.560) at \( \epsilon_N =0.01\) or \(0.05\). The Ellipsoid method consistently yields the second-lowest worst-case delays across the entire parameter range. By contrast, the SVC-based variants (e.g., SKL-based SVC with random grouping and MKL-based SVC) exhibit obvious sensitivity to parameter selection, with pronounces performance fluctuations. The extended SKL-based SVC with hierarchical grouping is comparatively more stable and achieves lower worst-case delays across a wide range of parameter settings.

To further examine performance across instance categories, we compute the average delay within each category. For each robust method, we select the parameter setting that yields the best performance under the high demand and high obstacle category. The results are illustrated in Table~\ref{tab:synthetic result1}. Under high-congestion conditions, all robust methods reduce average delay relative to the benchmark, whereas the performance gaps narrow under low-congestion conditions. The SKL-based SVC with random grouping achieves the lowest average delay in the high-demand and high-obstacle category, slightly outperforming the Ellipsoidal method. But its advantage does not consistently carry over to the remaining categories, where its performance becomes less competitive.  

\begin{table}[htbp]
\centering
\caption{Average delay of different methods across instance categories.}
\label{tab:synthetic result1}
\renewcommand{\arraystretch}{1.15}
\setlength{\tabcolsep}{6pt}
\begin{tabular}{c c c c c c c c c}
\hline
Demand &
Obstacle &
Benchmark & 
Budget & 
Ellipsoid & 
\makecell{SVC- \\ random} & 
\makecell{SVC- \\ hierarchical} & 
\makecell{MKL- \\ hierarchical} & 
DRSP \\

\hline
high   & high   & 0.601     & 0.485  & 0.448     & 0.446      & 0.462            & 0.452            & 0.452 \\
high   & medium & 0.583     & 0.478  & 0.468     & 0.486      & 0.499            & 0.464            & 0.469 \\
high   & low    & 0.572     & 0.445  & 0.441     & 0.441      & 0.453            & 0.439            & 0.445 \\
medium & high   & 0.423     & 0.356  & 0.356     & 0.362      & 0.364            & 0.357            & 0.356 \\
medium & medium & 0.431     & 0.365  & 0.361     & 0.352      & 0.343            & 0.361            & 0.355 \\
medium & low    & 0.419     & 0.354  & 0.361     & 0.363      & 0.361            & 0.361            & 0.359 \\
low    & high   & 0.218     & 0.245  & 0.235     & 0.237      & 0.233            & 0.248            & 0.234 \\
low    & medium & 0.207     & 0.226  & 0.223     & 0.226      & 0.221            & 0.228            & 0.227 \\
low    & low    & 0.244     & 0.229  & 0.239     & 0.23       & 0.232            & 0.224            & 0.241 \\
\hline
\end{tabular}
\end{table}

Overall, this synthetic experiment provides a controlled setting to analyze robust methods under systematically varied congestion and obstruction conditions. While this setting enables clear performance comparisons under well-defined conditions, it does not capture the spatial heterogeneity, demand variability, and network irregularities present in real urban environments. To examine whether the observed trends persist under realistic operating conditions, we next evaluate the same methods on a detailed pedestrian network in central Stockholm.

\subsection{Central Stockholm network}
\label{sec:scenario setup}
\begin{figure}
    \centering
    \begin{subfigure}{0.45\textwidth}
        \centering
        \includegraphics[width=\linewidth]{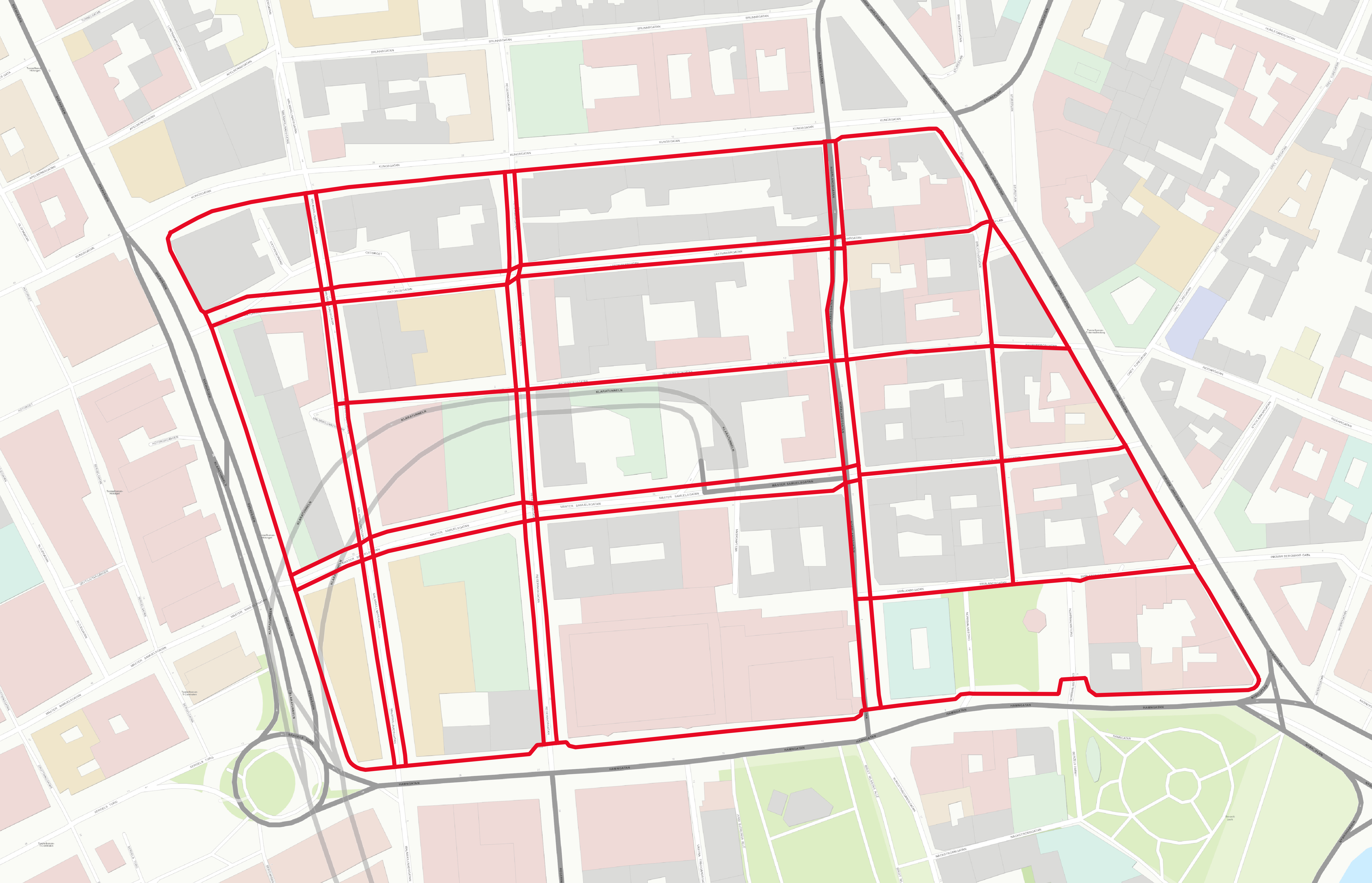}
        \caption{Network in QGIS}
        \label{fig:5a}
    \end{subfigure}
    \hfill
    \begin{subfigure}{0.52\textwidth}
        \centering
        \includegraphics[width=\linewidth]{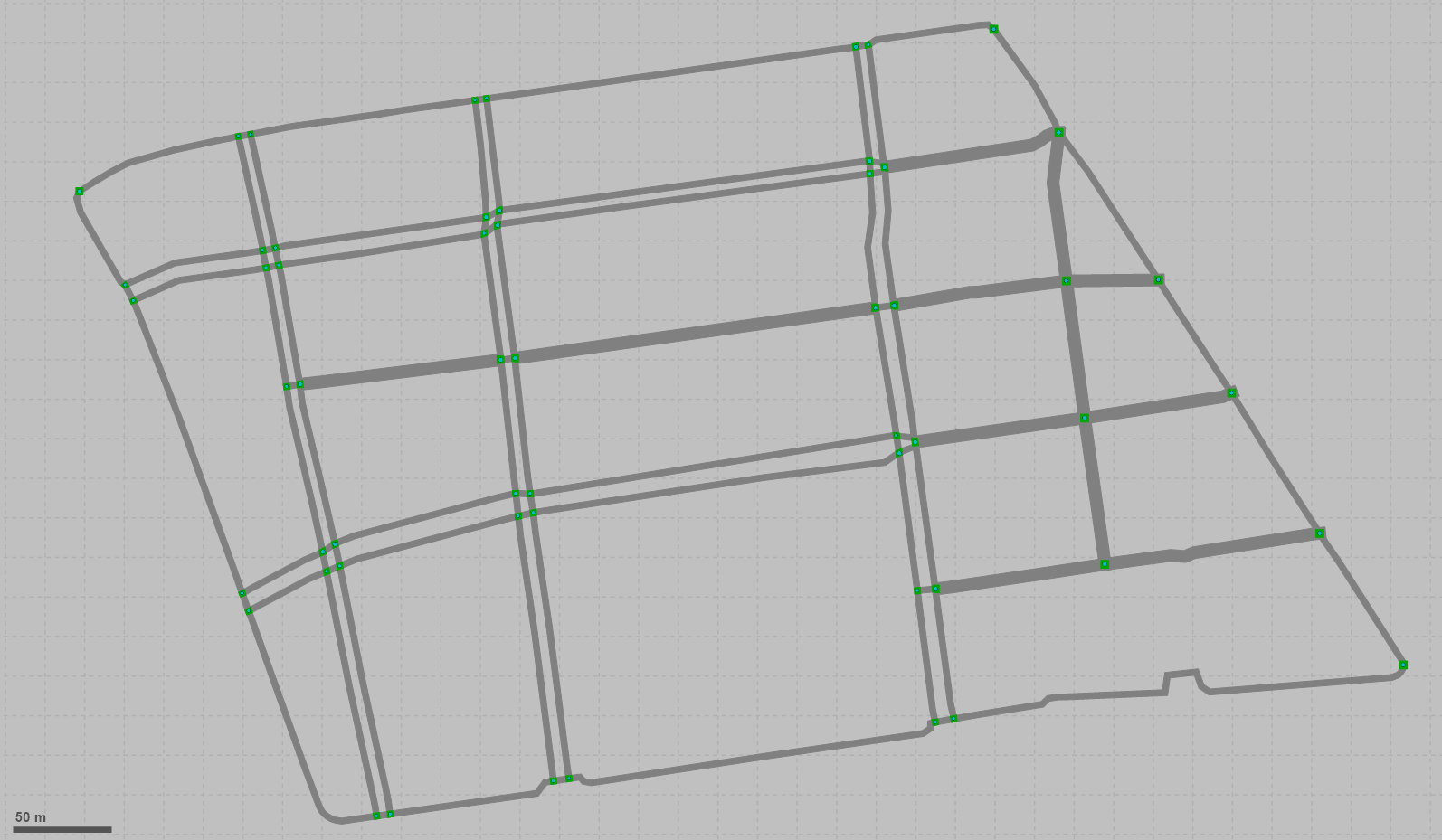}
        \caption{Network in VISWALK}
        \label{fig:5b}
    \end{subfigure}
    \caption{Refined Pedestrian network in Norrmalm, Stockholm}
    \label{fig:5}
\end{figure}

Building on the insights obtained from the synthetic network, we now consider a realistic case study based on a section of Norrmalm, located in central Stockholm, bounded by Kungsgatan, Birger Jarlsgatan, Hamngatan, and Sveavägen. This area was designated as a Class 3 clean air zone which allows only fully electric vehicles. It is also characterized by high pedestrian traffic and complex interactions between obstacles and pedestrians, making it a particularly relevant setting for studying sidewalk robot delivery. The network geometry was obtained from OpenStreetMap and refined using QGIS to ensure accuracy (Figure~\ref{fig:5}). The network has 99 segments, including 65 sidewalks and 34 crossings. Note that this network size is particularly appropriate for sidewalk delivery robots, since they are designed for last-mile logistics and generally do not undertake long-distance routes. It is both realistic and practical to model a compact, urban network rather than a large-scale city-wide system.

The VISWALK simulation input includes road network geometry, pedestrian demand, behavioral parameters, and obstacles, all calibrated to reflect real-world pedestrian and robot behavior. The simulation duration is set as 900 seconds (15 minutes). Pedestrian origin-destination (OD) matrices are constructed for each hour from 10 a.m. to 10 p.m. (12 hours) across 7 days of the week. These are combined with 12 distinct obstacle configurations and bidirectional travel for each segment, resulting in a total of 2016 simulation scenarios. 

\begin{figure}
    \centering
    \begin{subfigure}{0.46\textwidth}
        \centering
        \includegraphics[width=\linewidth]{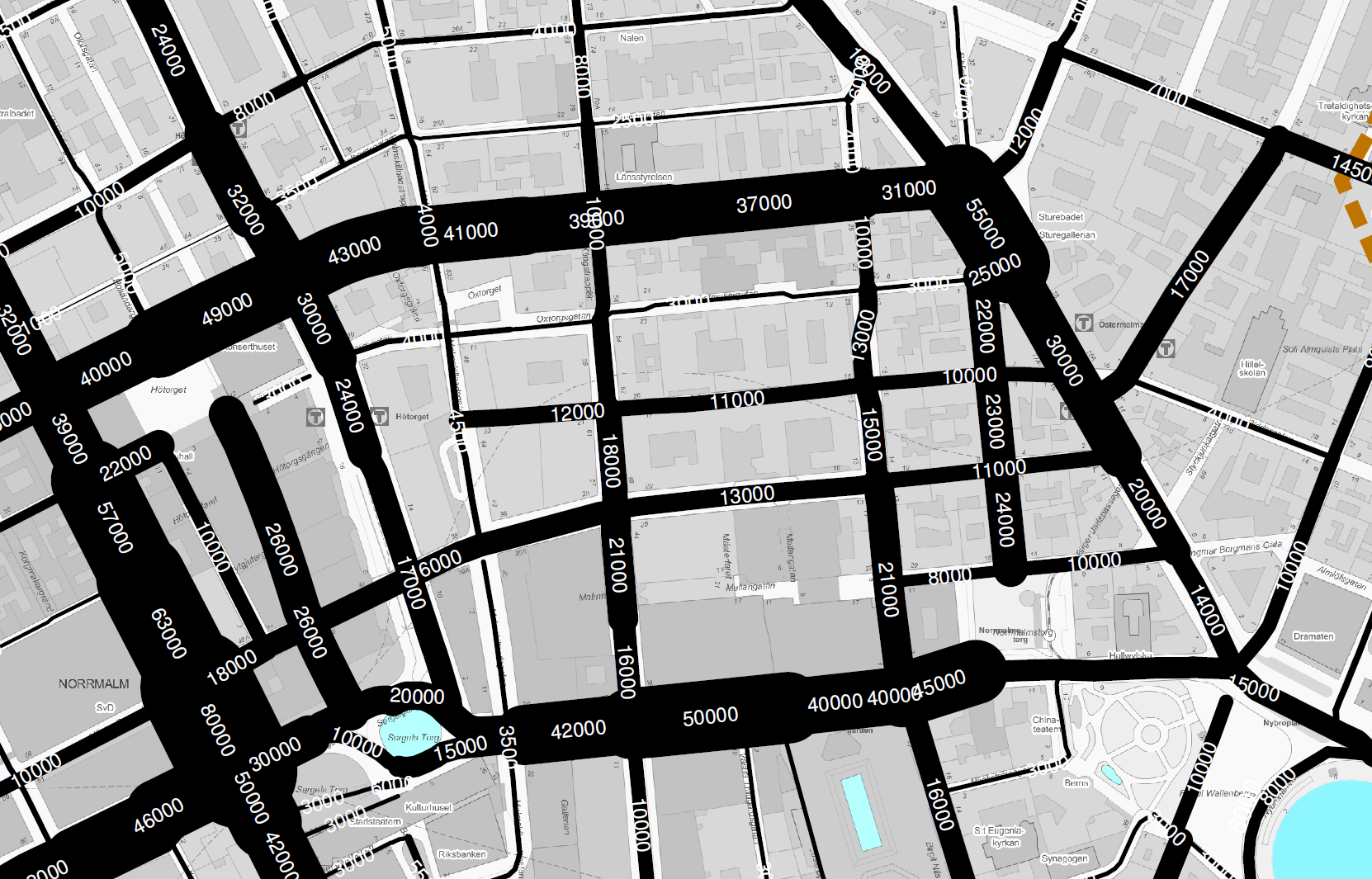}
        \caption{Average Daily Pedestrian flow of central Stockholm}
        \label{fig:6a}
    \end{subfigure}
    \hfill
    \begin{subfigure}{0.52\textwidth}
        \centering
        \includegraphics[width=\linewidth]{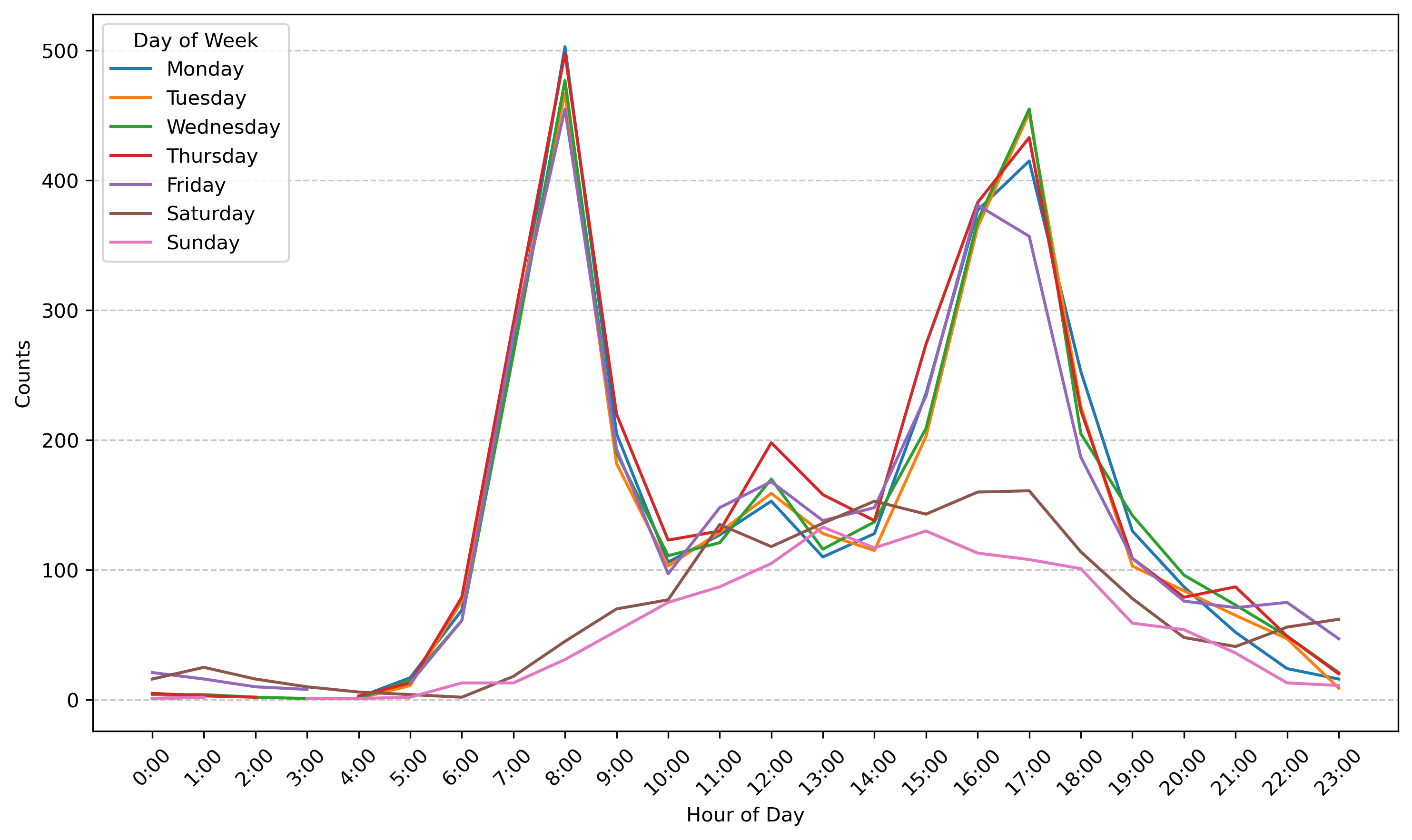}
        \caption{Distribution of Pedestrian volume across different days of the week and hours of the day in central Stockholm}
        \label{fig:6b}
    \end{subfigure}
    \caption{Sources of pedestrian demand data}
    \label{fig:6}
\end{figure}

The pedestrian daily traffic volumes data were obtained from openly available data shared by Stockholm municipality \citep{stlmPedFlow}. This resource provides a detailed flow map showing the average number of pedestrians per day on streets in central Stockholm during the year 2017 (Figure~\ref{fig:6a}). To simulate realistic pedestrian flows over time, these volumes were further refined based on the distribution of pedestrian activity across different days of the week and hours of the day. This approach utilized data from a 2015 Stockholm regional travel behavior survey \citep{stlm2015}, which includes full-day travel schedules of approximately 11,500 individuals in Stockholm County. Specifically, we focused on trips originating from or directed to the central Stockholm area to derive this distribution (Figure~\ref{fig:6b}). Based on this distribution, the pedestrian volume for a specific day and time was calculated, and then the corresponding 15-minute flow rates were derived for each segment, enabling the construction of OD matrices for the simulation. The pedestrian population follows the standard composition and features: half are men with speeds uniformly distributed between 3.49–5.83 km/h, and half are women with speeds between 2.56–4.28 km/h.

\begin{figure}[t]
    \centering
        \includegraphics[width=0.7\linewidth]{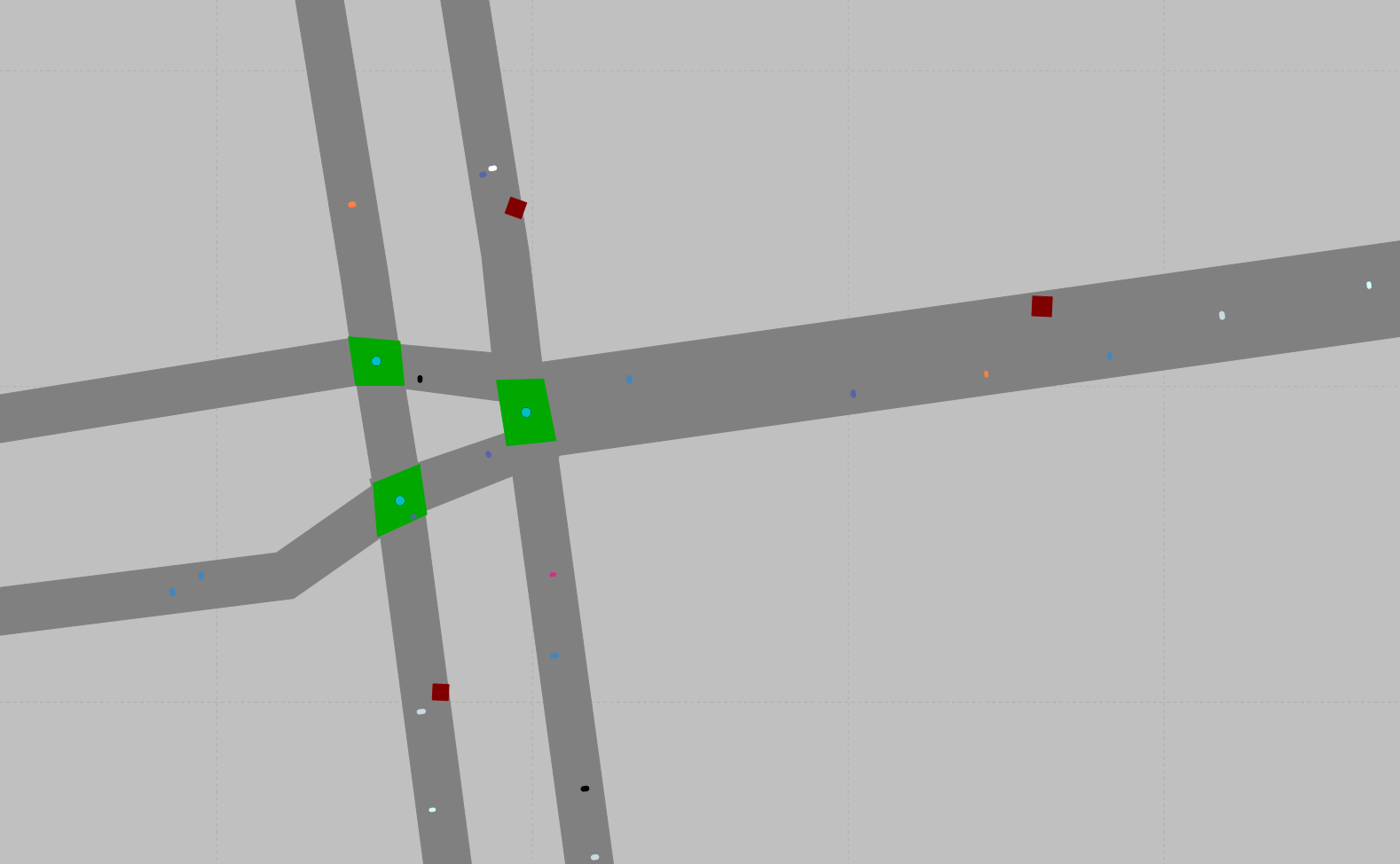}
    \caption{Obstacles (red squares) on the sidewalk network }
    \label{fig:7}
\end{figure}

Obstacles in the simulation play a key role in influencing pedestrian movement patterns. In this study, we model various obstacles to reflect real-world barriers and elements. The obstacle placement is based on the actual street layouts, more specifically the number of sidewalks. Given a percentage, a corresponding share of road segments is selected, and an obstacle of random size is placed at a random location on each. We set up 12 obstacle configurations with the following percentages: 0, 10\%, 20\%, 30\%, 40\%, 50\% (each repeated twice). The obstacles generated are squares with side lengths randomly varying between 1 and 1.4 meters. Figure~\ref{fig:7} illustrates an example of obstacles placed within the sidewalk network.

\subsection{Performance analysis}
\label{sec:performance analysis}
To assess the need of RSP algorithms over conventional shortest path algorithms, we first analyze how optimal paths vary across various simulation scenarios. The conventional shortest paths are computed for 500 OD pairs, ensuring each path includes at least five segments. Results show that only five OD pairs maintain the same path across all scenarios, while 346 have at least 10 different paths, and 33 exhibit over 50 different paths. On average, each OD pair has 19.25 unique optimal paths, highlighting the need for robust shortest path approaches. 

\begin{figure}
    \centering
    \begin{subfigure}{0.48\textwidth}
        \centering
        \includegraphics[width=\linewidth]{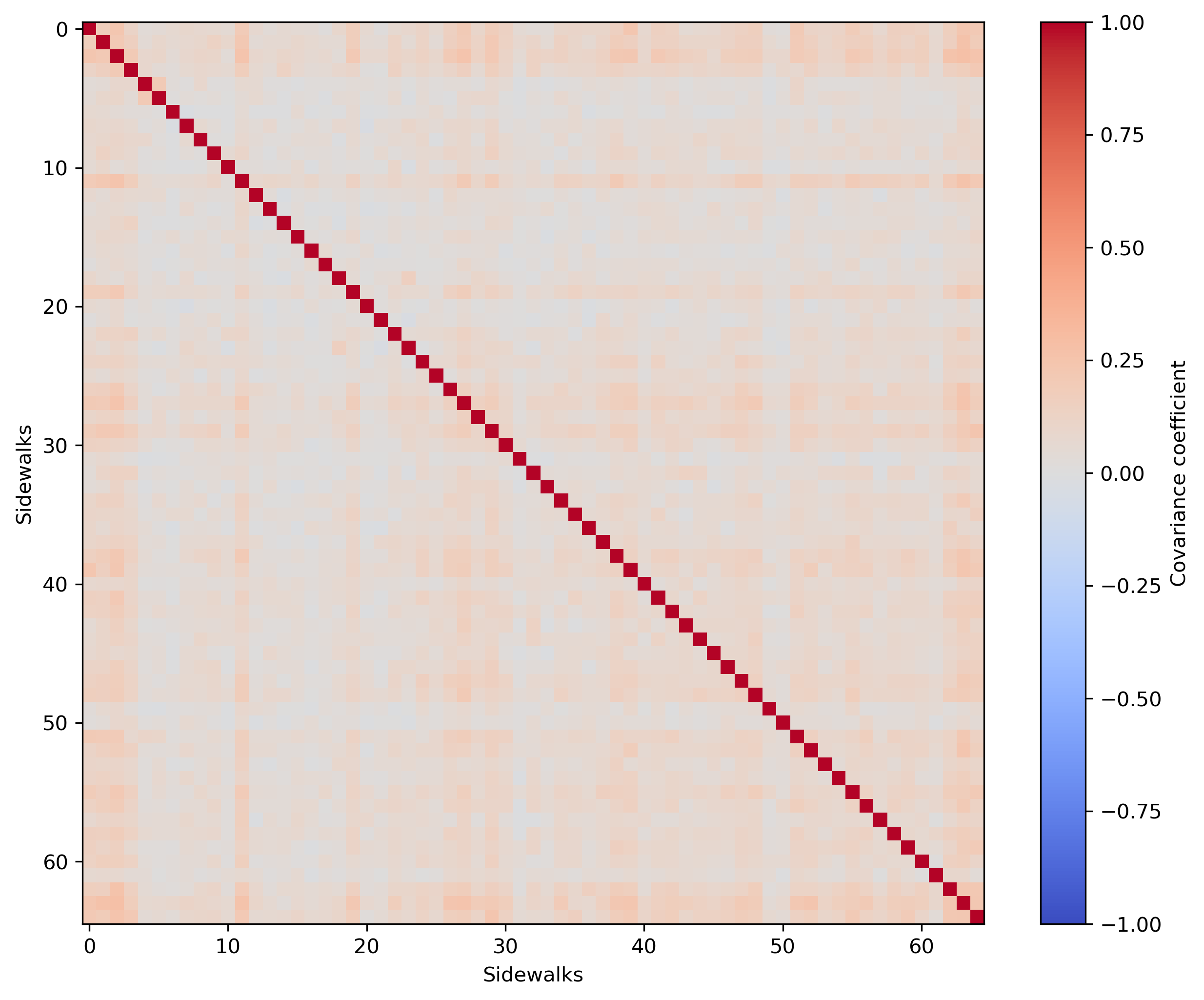}
        \caption{Correlation between sidewalks}
        \label{fig:8a}
    \end{subfigure}
    \hfill
    \begin{subfigure}{0.48\textwidth}
        \centering
        \includegraphics[width=\linewidth]{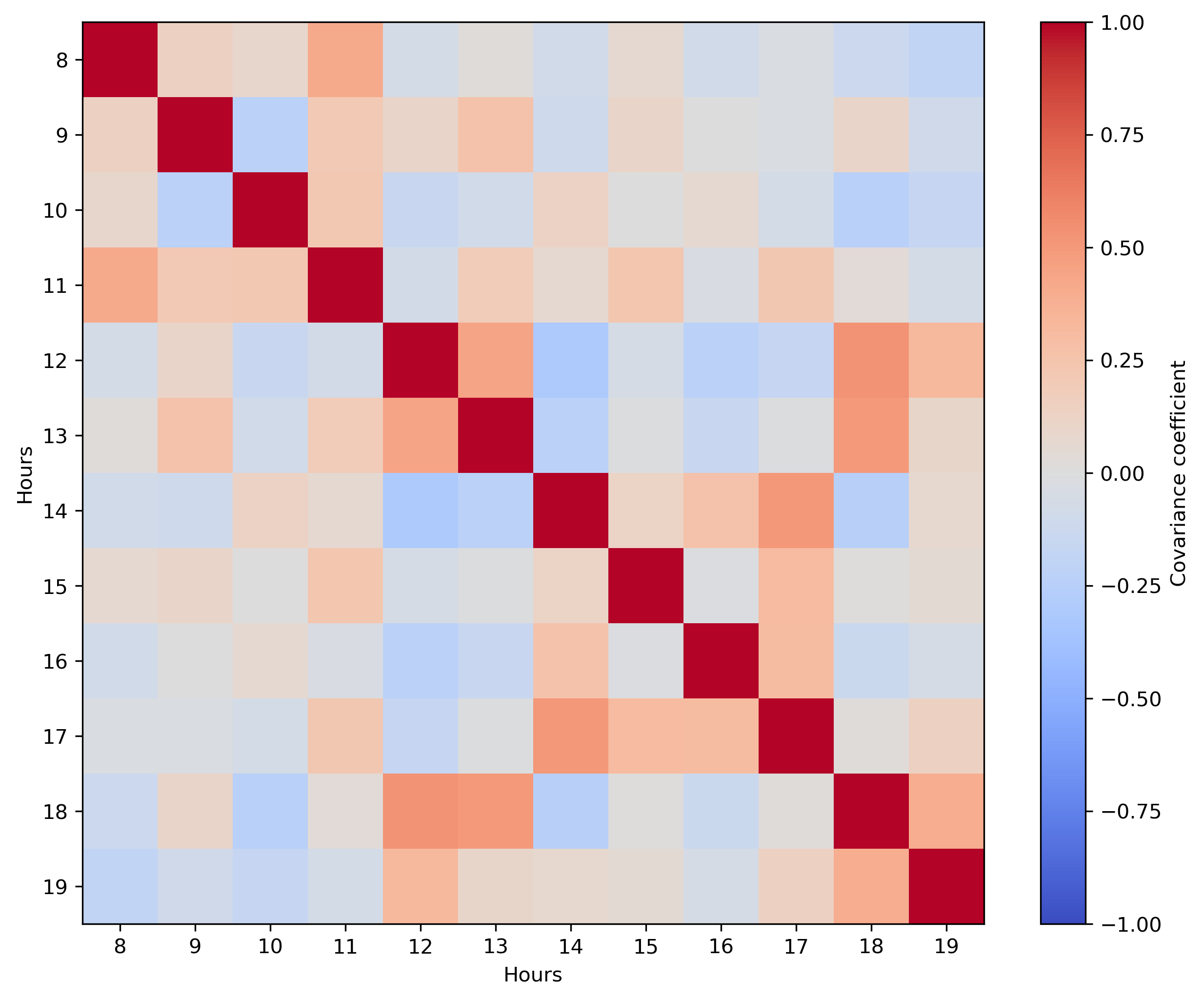}
        \caption{Correlation between hours}
        \label{fig:8b}
    \end{subfigure}
    \caption{Correlation between segments and between hours}
    \label{fig:8}
\end{figure}

Before generating uncertainty (or ambiguity) sets, we analyze travel time correlations among segments and across different hours. Figure~\ref{fig:8a} illustrates the correlation coefficients for each pair of sidewalks. Notably, a weak positive correlation exists among certain pairs of sidewalks. However, the absolute value of the correlation coefficient for 90\% of the pairs does not exceed 0.14, and the highest value is only 0.27, indicating a near absence of significant correlation between sidewalks, implying that travel time fluctuations on one sidewalk are relatively independent of those on others. The correlation across various hours is shown in Figure~\ref{fig:8b}. We aggregate the average travel time over all segments in the network and derive the temporal correlation between different hours. Variations across hours are driven by differences in simulated pedestrian volumes. The results indicate relatively strong temporal correlations, with 10 out of 66 hour pairs exhibiting absolute correlation values above 0.3, and two pairs exceeding 0.5. High correlation during certain hours suggests consistent and predictable travel time patterns, implying reduced temporal uncertainty. While such predictability could benefit the implementation of time-dependent algorithms, we focus here on a time-independent approach for better tractability and transferability. 

Based on the low spatial correlation, it is possible to apply the TSC-DS algorithm for SVC, which resolves the over-conservatism caused by high-dimensional data and improves the computational tractability (see Section \ref{sec:svc}). The 99 segments are randomly grouped into 49 paired two-dimensional subsets (one subset has three segments), effectively reducing the dimensionality from 99 to 2. While the standard SVC based on 1612 simulated data points would typically yield over 322 SVs, the proposed algorithm reduces such a number to just over 12 SVs. As a result, the average running time for a single origin-destination pair decreases from 11,552 seconds to 13.8 seconds, on a system equipped with an Intel Core i7-1365U processor. 

The parameter ranges of robust methods are selected based on the same parameter definitions and algorithmic considerations as in the synthetic experiment. The fixed parameters for the MKL-based SVC and DRSP methods are selected based on preliminary experiments. For the MKL-based SVC method, we adopt \(M=16\) basis kernels and set \(\mu=0.2\) based on sensitivity analysis over the range \(\mu \in [0.05, 0.5]\), which showed limited variation in results and justified using a fixed value. The DRSP approach involves three parameters: \(\alpha\), \(\epsilon_N\), and N. To simplify the comparison and reduce dimensionality, we vary each parameter individually while holding the others constant and identify stable values through preliminary pruning. As a result, we fix \(\alpha=0.3\) and \(N=500\), and vary \(\epsilon_N\) in the main analysis. 

Similar to the synthetic experiment, we compute the free-flow travel time of each segment as the ratio of segment length to the robot's desired speed, and use the resulting conventional shortest path as a benchmark for the various robust approaches. A five-fold validation is also performed. In each fold, 1612 (80\%) of the 2016 scenarios are used for training, and 404 (20\%) for validation, sampled from the travel time matrix. From the initial 500 OD pairs, the 100 with the highest travel time standard deviation are selected, resulting in 1,656,400 travel cost calculations per fold. We report three performance criteria proposed by \citet{chassein2018}: 
\begin{itemize}
    \item the average travel time across all OD pairs and scenarios, 
    \item the average worst-case travel time across scenarios for all OD pairs, 
    \item the average of the 5th percentile worst travel times across all scenarios for all OD pairs.
\end{itemize}
For comparability across OD pairs with different distances, the travel time of each OD pair is normalized by the free flow travel time of the benchmark path. Accordingly, the three performance criteria are referred to as average delay, worst-case delay, and worst 5\% delay, respectively.

\begin{figure}[t]
    \centering
    \begin{subfigure}{0.7\textwidth}
        \centering
        \includegraphics[width=\linewidth]{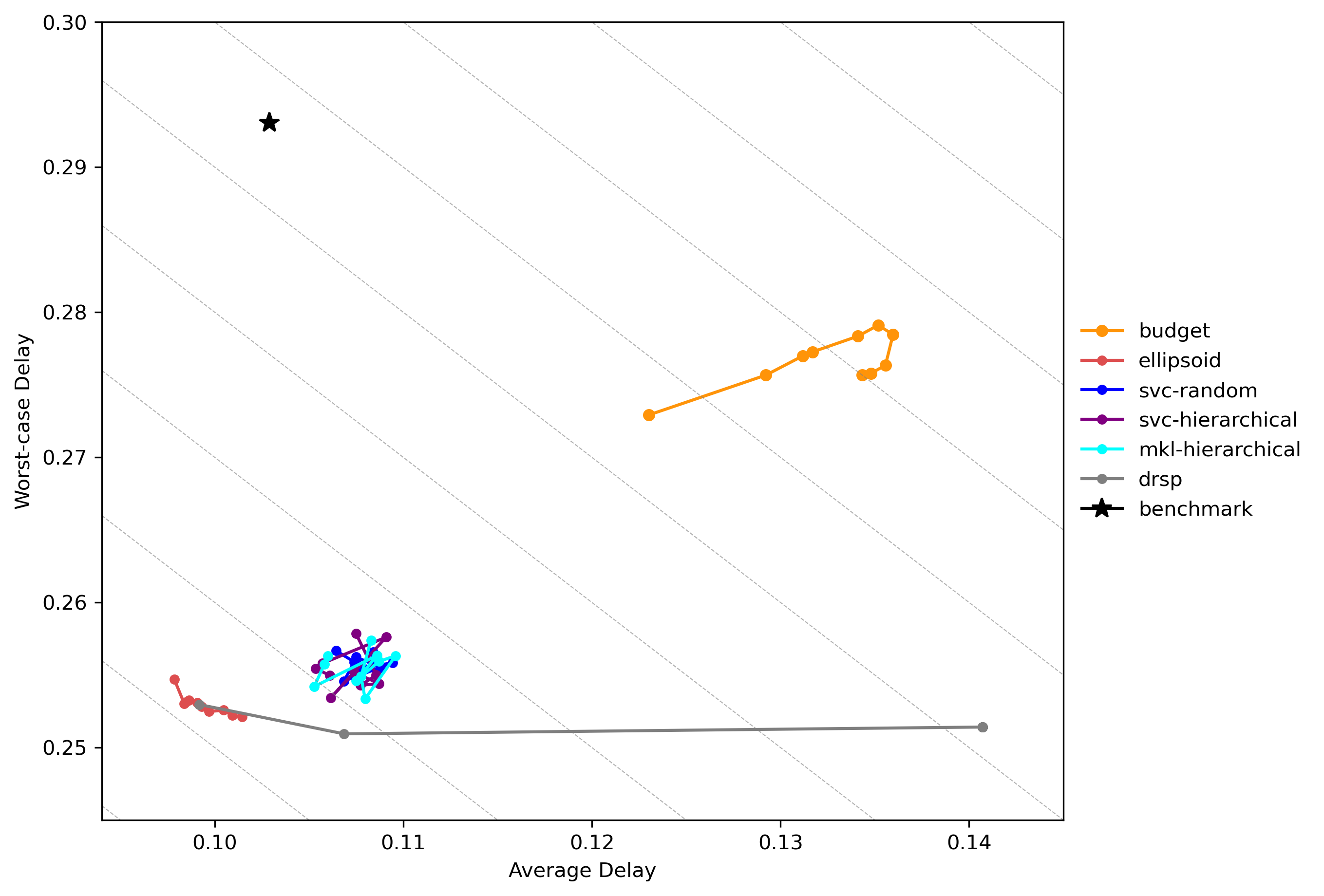}
        \caption{Average vs. Worst case }
        \label{fig:9a}
    \end{subfigure}

    \begin{subfigure}{0.7\textwidth}
        \centering
        \includegraphics[width=\linewidth]{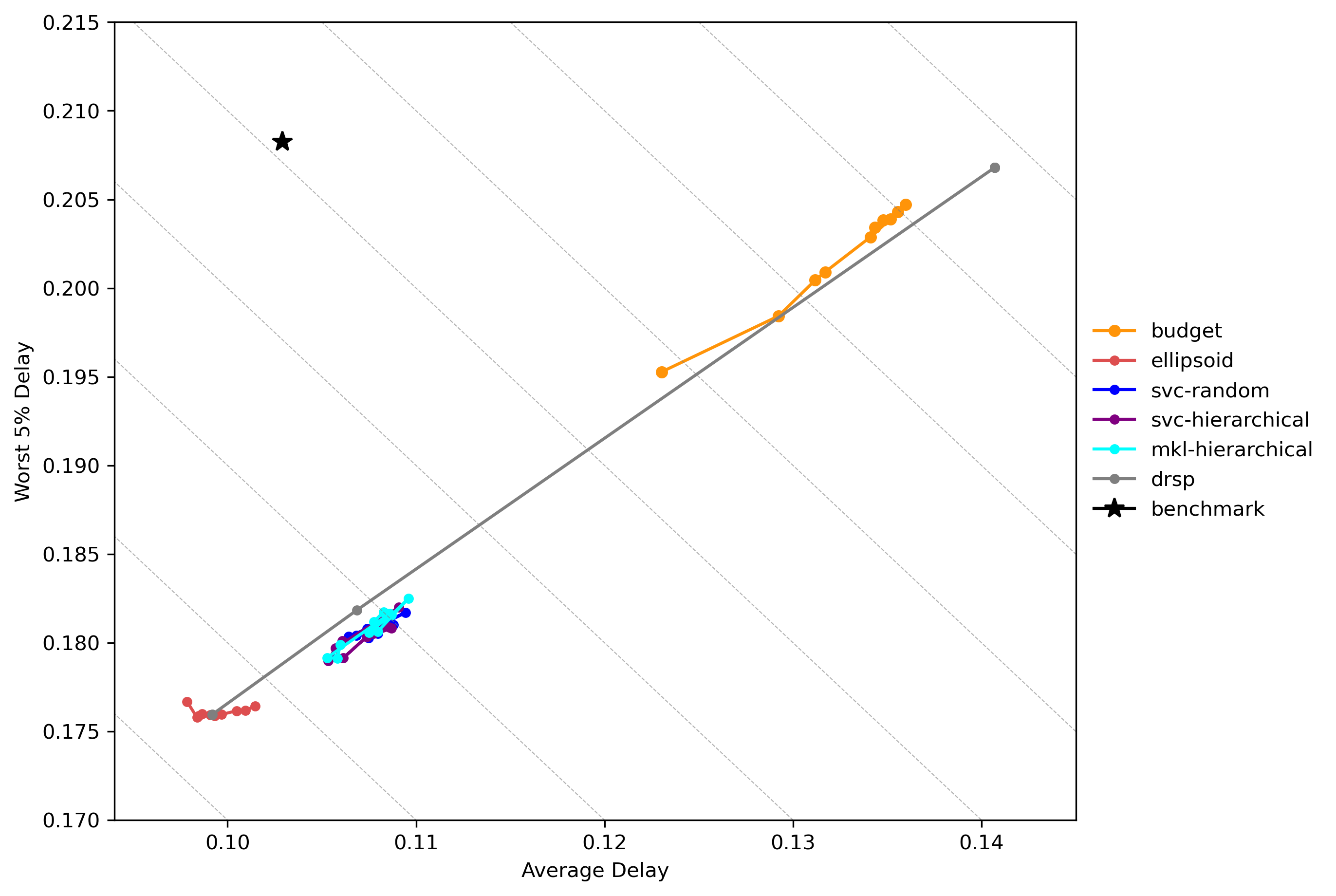}
        \caption{Average vs. Worst 5\%}
        \label{fig:9b}
    \end{subfigure}
    \caption{Trade-off between three performance criteria}
    \label{fig:9}
\end{figure}

The performance of various methods under different parameter settings is illustrated in Figure~\ref{fig:9}. Robust paths generally incur slightly higher average delay than the conventional shortest path, as they explicitly hedge against uncertainty and worst-case travel conditions, but they achieve lower delays in adverse scenarios. Lower values on both axes indicate better performance; hence, points closer to the lower-left corner represent more desirable trade-offs between efficiency and robustness. To facilitate visual comparison of this trade-off, grid lines with a slope of -1 are included in the figures.

Figure~\ref{fig:9a} compares average delay and worst-case delay across different methods and parameter settings. All robust methods yield lower worst-case delays than the benchmark, which has an average worst-case delay of 0.293. Under the Budgeted uncertainty set, the best result is achieved with \( \Gamma =1\), reducing the worst-case delay by up to 6.9\%. However, increasing \( \Gamma\) leads to overly conservative solutions characterized by poorer performance.

The Ellipsoidal, SKL- and MKL-based SVC, and DRSP perform substantially better. For Ellipsoidal uncertainty, the lowest worst-case delay (14\% reduction) is achieved at \( \lambda =10\), but it comes at the cost of the largest average delay. Smaller \( \lambda\) values lead to higher worst-case delays but lower average delays. Notably, the Ellipsoidal method consistently achieves lower average delays across all parameters than the conventional SP, with reductions of up to 4.9\%, indicating higher routing efficiency.

The DRSP method achieves the overall lowest worst-case delay (0.251) at \( \epsilon_N =0.1\), but its average delay increases markedly beyond this parameter. For the SVC-based methods, all three variants show comparable overall performance. However, the SKL- and MKL-based extensions, which incorporate hierarchical clustering, outperform the original SVC with random grouping under specific parameter settings (\( v =0.12\) and \( v =0.16\), respectively). They reach maximum reductions in worst-case delay of 13.5\% and 13.3\% relative to the benchmark.

When analyzing performance for the worst 5\% travel delays (Figure~\ref{fig:9b}), the differences between parameter settings are more pronounced. Again, the Ellipsoidal method performs consistently better than the other methods. The DRSP method performs well, but its performance deteriorates sharply as \( \epsilon_N\) increases, indicating a high sensitivity to parameter selection. Although the SVC-based methods underperform compared to the Ellipsoidal and DRSP approaches, the hierarchical extension shows a more pronounced improvement over the original SVC in reducing the worst 5\% of delays.  

\begin{figure}[t]
    \centering
    \begin{subfigure}{0.49\textwidth}
        \centering
        \includegraphics[width=\linewidth]{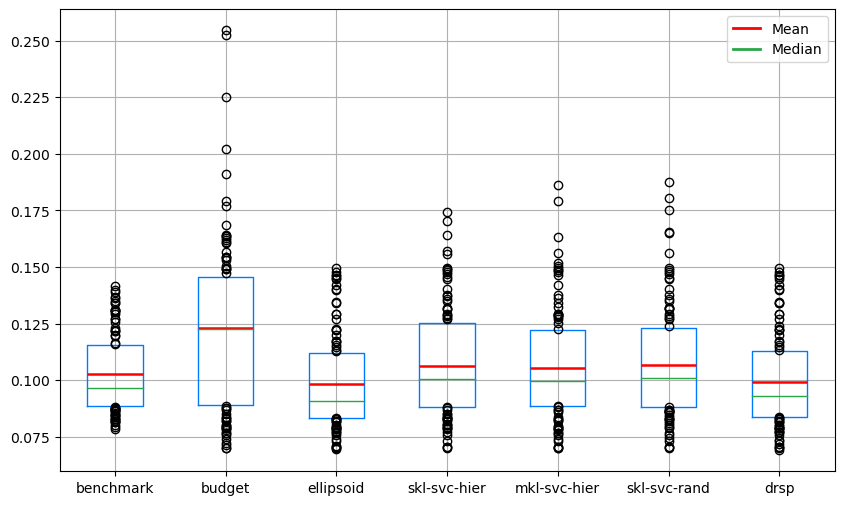}
        \caption{Average delay of different OD pairs}
        \label{fig:10a}
    \end{subfigure}
    \hfill
    \begin{subfigure}{0.49\textwidth}
        \centering
        \includegraphics[width=\linewidth]{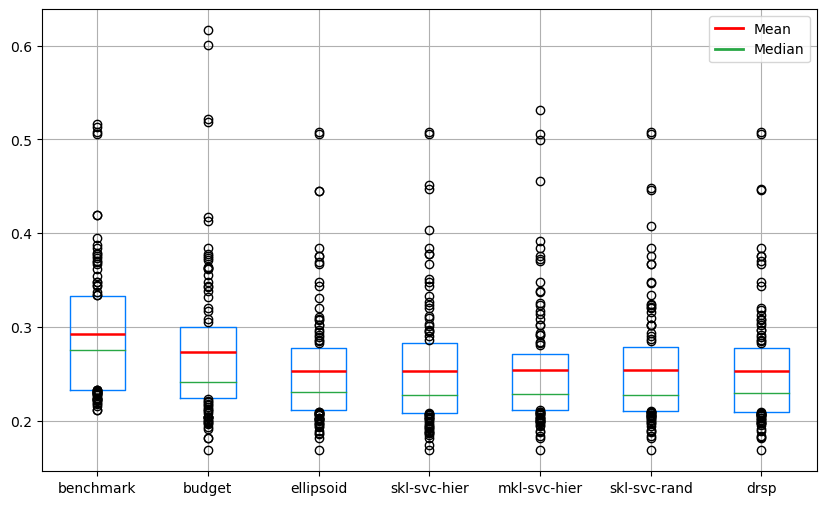}
        \caption{Worst delay of different OD pairs}
        \label{fig:10b}
    \end{subfigure}
    \begin{subfigure}{0.51\textwidth}
        \centering
        \includegraphics[width=\linewidth]{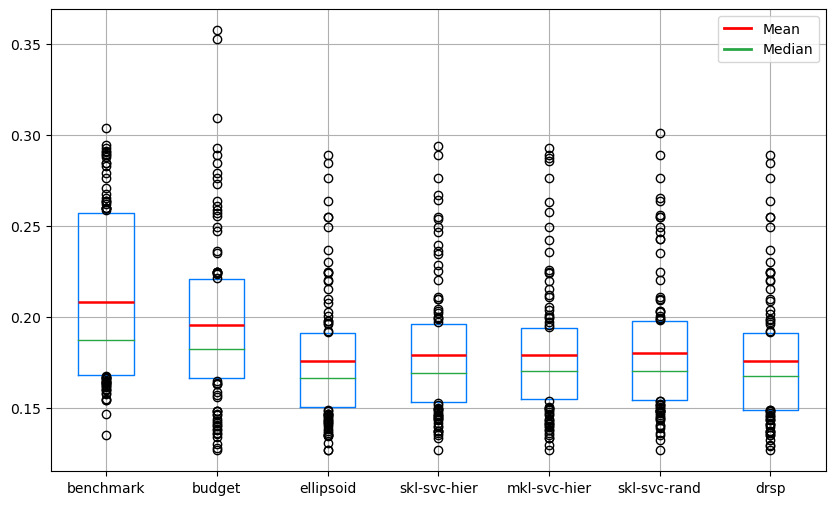}
        \caption{Worst 5\% delay of different OD pairs}
        \label{fig:10c}
    \end{subfigure}
    \caption{Distribution of three types of uncertainty sets for different OD pairs}
    \label{fig:10}
\end{figure}

We integrate the analysis of aggregated results with a more detailed analysis of three performance metrics across individual OD pairs. For a fair comparison, the best-performing parameters for each robust method are selected based on the previous analysis: \( \Gamma =1 \) for Budgeted, \(\lambda =3 \) for Ellipsoidal,  \(v =0.12 \) for SKL-based SVC, \(v =0.16 \) for MKL-based SVC, \(v =0.16 \) for original SVC, and \( \epsilon_N =0.05\) for DRSP. The results illustrated in Figure~\ref{fig:10} as box plots, with lower and more compact boxes indicating higher efficiency and robustness.

Figure~\ref{fig:10a} shows the distribution of average travel delays. All robust methods display greater variability compared to the conventional SP method. Among them, the Ellipsoidal method achieves lower and more concentrated average delays than the other robust approaches. All robust methods outperform the conventional approach in terms of worst-case delay variability (Figure~\ref{fig:10b}), with distributions that are both narrower and lower. The Ellipsoidal method stands out with a slightly lower distribution and the lowest average worst-case delay. The advantages of robust methods are even more evident in the worst 5\% scenarios, illustrated in Figure~\ref{fig:10c}. Both the Ellipsoidal and DRSP methods exhibit significantly lower medians and means, along with tighter distributions, indicating strong performance in the most adverse conditions. Although the Budgeted method demonstrates improved robustness over the benchmark for most OD pairs, it performs even worse than the benchmark for certain individual OD pairs, which is not observed with the other robust methods.

These findings indicate that well-tuned robust approaches significantly reduce worst-case delays across OD pairs with high uncertainty, compared to the traditional shortest path method. Among them, the Ellipsoidal and DRSP methods not only outperform other robust methods in terms of worst-case delays, but also achieve lower average delays than the conventional approach. While the Ellipsoidal method performs consistently well across parameters, the DRSP method demonstrates strong performance when properly tuned, highlighting its potential under careful calibration. In addition to strong average performance, their consistent results across individual OD pairs highlight greater stability and practical reliability. The adopted five-fold validation approach guarantees these results can be generalized without overfitting risks.

In contrast with our expectations, the data-driven SVC-based approaches, while outperforming the Budgeted method in terms of robustness, fail to surpass the Ellipsoidal method in either robustness or efficiency. This occurs regardless of the investigated SVC variant (single or multi-kernel). When considering computational cost, this performance gap becomes even more pronounced since the scalability of SVC becomes problematic for large problem sizes. Moreover, the marginal improvement achieved by incorporating hierarchical clustering suggests that addressing weak inter-segment relationships in SVC does not yield substantial gains and is insufficient to outperform the Ellipsoidal or DRSP methods. 

One possible explanation lies in the characteristics of the dataset itself. The input data consists of a simulated sidewalk robot travel time matrix which may be relatively ``symmetric". Its distribution may be well approximated by an ellipsoid fitted to the observed data, rendering the added complexity of SVC methods unnecessary. Additionally, data-driven methods typically require larger datasets to effectively learn complex patterns. For applications with more limited data, the Ellipsoid method seems to be more suitable. Further investigation is needed to better understand the impact of dataset size and structure on method performance. 

\subsection{Sensitivity analysis}
\label{sec: sensitivity_analysis}

In real-world sidewalk robot operations, various design and environmental factors can influence robot performance (respectively discussed in Section~\ref{sec: robot design features} and Section~\ref{sec: environmental factors}) can affect the comparative advantage of robust methods over the conventional SP approach. To investigate this, a one-way local sensitivity analysis is conducted to evaluate how variations in robot design features and environmental factors impact the performance of robust methods.

Each factor is varied independently across three configurations, which include a default case aligned with the main experiments in Section~\ref{sec:performance analysis}, except for obstacle placement, which is adjusted to the sidewalk edges to avoid unrealistic blockages and slightly reduce travel time variability.

Each case is simulated under 2016 scenarios as before (see Section~\ref{sec:scenario setup}). The resulting travel time matrices are fed into the same performance evaluation framework described in Section~\ref{sec:performance analysis}, consisting of the following KPIs: average delays, worst-case delays, and worst 5\% delays. To clearly illustrate the improvement of robust methods over the SP approach, we report the difference in each performance criterion between the SP and robust methods, where higher values indicate greater delay reduction and, thus, better performance. 

Since our previous analyses identified the Ellipsoidal and DRSP methods as the most promising robust approaches, we limit the sensitivity analysis to these two. We re-evaluate them using the same parameter tuning ranges as in the main experiments:
\begin{itemize}
    \item Ellipsoidal uncertainty: \( \lambda \in \{1,2,...,10\}\);
    \item DRSP: \( \epsilon_N\in \{0.01,0.05,0.1,0.2,...,0.8\}\), with fixed \(\alpha=0.3\) and \(N=500\). 
\end{itemize}

\subsubsection{Impacts of robot design features}
\label{sec: robot design features}

Key robot design features, including desired speed, width, and navigation behavior, can affect the relative performance of robust methods compared to the SP approach. We analyze the trade-offs between improvements in average and worst-case travel delays across different robot design configurations. 

\begin{figure}[t]
    \centering
    \begin{subfigure}{0.49\textwidth}
        \centering
        \includegraphics[width=\linewidth]{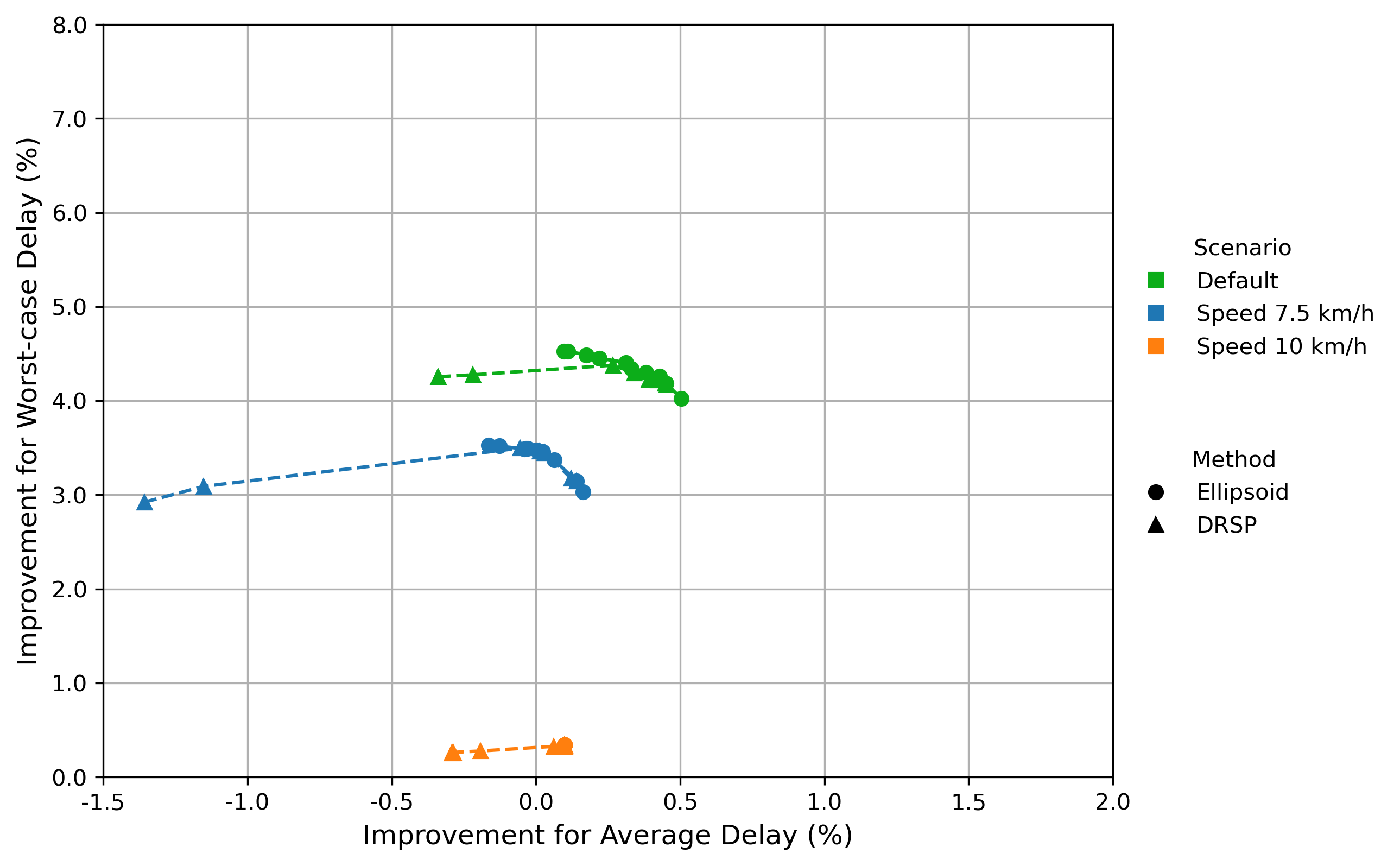}
        \caption{Robot desired speed}
        \label{fig:11a}
    \end{subfigure}
    \hfill
    \begin{subfigure}{0.49\textwidth}
        \centering
        \includegraphics[width=\linewidth]{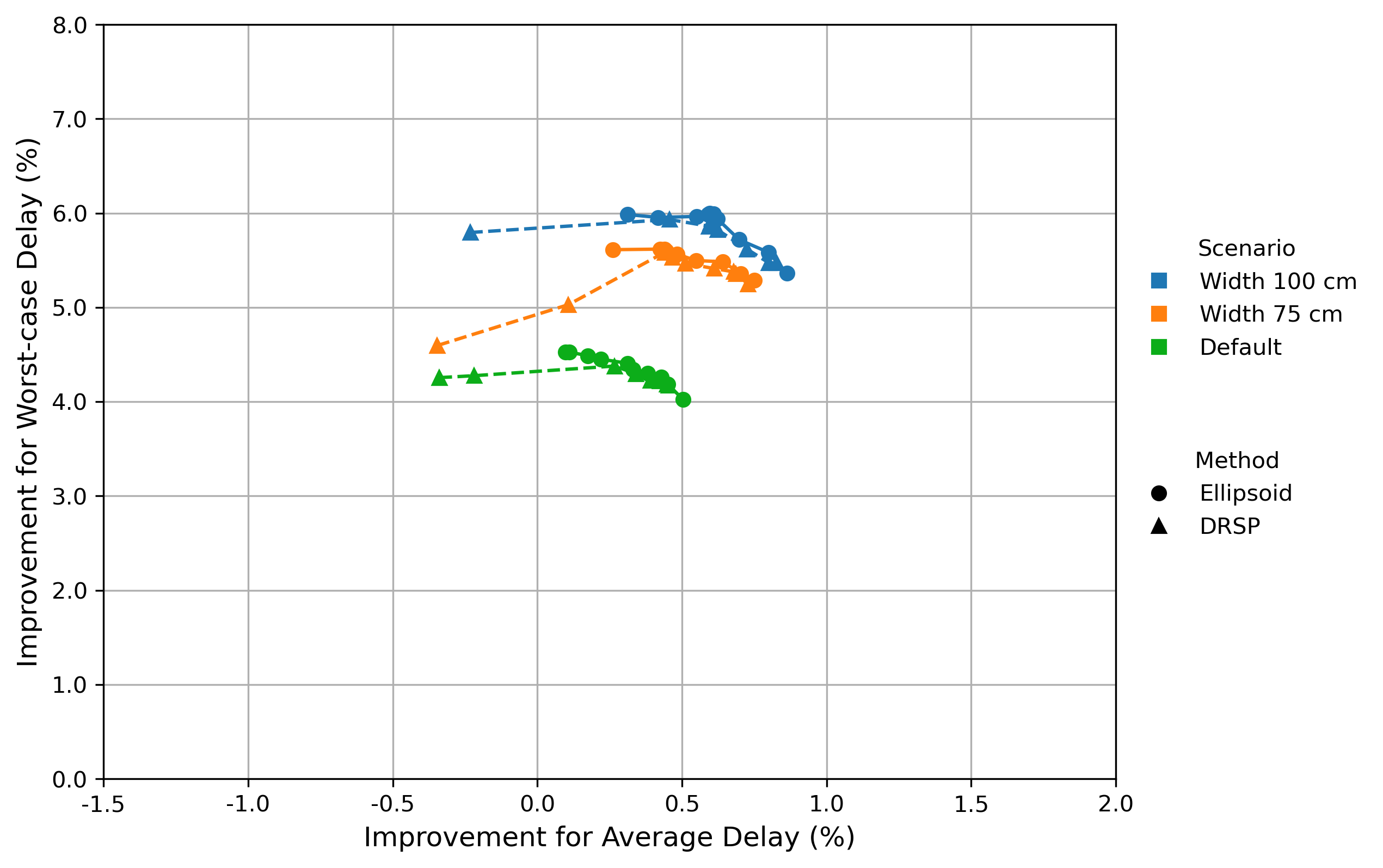}
        \caption{Robot width}
        \label{fig:11b}
    \end{subfigure}
    \begin{subfigure}{0.51\textwidth}
        \centering
        \includegraphics[width=\linewidth]{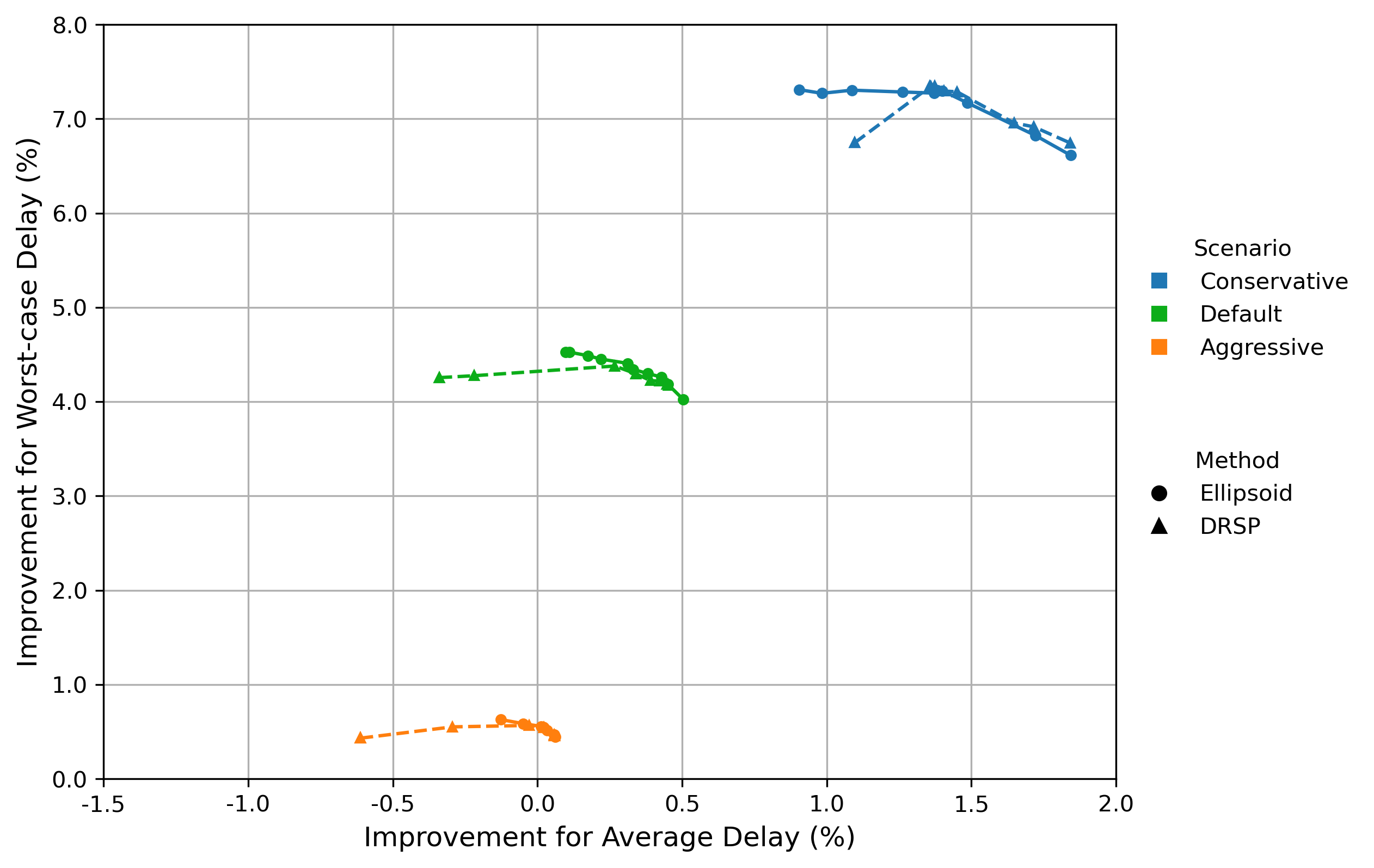}
        \caption{Robot behavior}
        \label{fig:11c}
    \end{subfigure}
    \caption{Performance improvement of selected robust methods under various robot design conditions}
    \label{fig:11}
\end{figure}

Three distinct levels of robot desired speed are considered in Figure~\ref{fig:11a}: 5 km/h (default), 7.5 km/h, and 10 km/h, representing a range from slower to relatively faster-moving robots. As the desired speed increases, the benefits of robust methods diminish. Specifically, the maximum improvement in worst-case delay drops from 4.5\% at 5 km/h to just 0.3\% at 10 km/h, while the gain in average delay decreases from 0.5\% to only 0.1\% over the same range. This suggests that faster-moving robots experience lower variability in travel time, reducing the relative advantage of incorporating robustness. 

In Figure~\ref{fig:11b}, we evaluate robot widths of 54 cm (default), 75 cm, and 100 cm, covering the range from narrow to wide designs. The results show a clear performance increase, with worst-case delay improvements increasing from 4.5\% (50 cm) to 6.0\% (100 cm). Average delay reductions also improve slightly. Although wider robots may face greater difficulty navigating narrow or obstructed sidewalks, robust models demonstrate a stronger ability to manage this added uncertainty than the SP method. 

Robot navigation behavior is categorized into three profiles: conservative, default, and aggressive, each modeled using the SFM with distinct parameter settings listed in Table~\ref{tab:2}. Figure~\ref{fig:11c} shows that conservative robots gain the most from robust planning. The maximum improvement in worst-case delay reaches 7.4\% under conservative behavior, compared to just 0.6\% for aggressive behavior. Similarly, the maximum improvement in average delay is 1.8\% for conservative behavior, while only 0.06\% for aggressive behavior. This result suggests that robust methods are especially beneficial when robot behavior leads to increased interaction with pedestrians or greater routing variability. 

\begin{table}[t]
        \centering
        \caption{Social force model parameters for different robot behaviors}
        \footnotesize
        \renewcommand{\arraystretch}{1.2} 
        \begin{tabular}{|c|c|c|c|c|}
        \hline  
        \textbf{Parameters}& 
        $\boldsymbol{\tau}$ &
        $\boldsymbol{\lambda}$ &
        $\boldsymbol{A}_{\text{soc\_iso}}$ & 
        $\boldsymbol{B}_{\text{soc\_iso}}$  \\  
        \hline 
        \textbf{Conservative}& 1.2&  0.45& 3& 0.5\\ \hline 
        \textbf{Default}& 0.8&  0.45& 2& 0.35\\ \hline
 \textbf{Aggressive}& 0.4& 0.45& 1& 0.2\\\hline 
        \end{tabular}
    \label{tab:2}
    \end{table}

Overall, the robust approaches show higher reliability and efficiency than the SP model for most robot design settings investigated. These advantages are more pronounced for slower, wider, and more conservative robots.

Across all the robot design configurations, the Ellipsoidal method generally performs better than the DRSP method, with only occasional overlaps. Moreover, its lower performance variation suggests that ellipsoidal uncertainty set-based RSP may have a greater stability and easier parameter tuning. 

\subsubsection{Impacts of environmental factors}
\label{sec: environmental factors}

\begin{figure}[t]
    \centering
    \begin{subfigure}{0.6\textwidth}
        \centering
        \includegraphics[width=\linewidth]{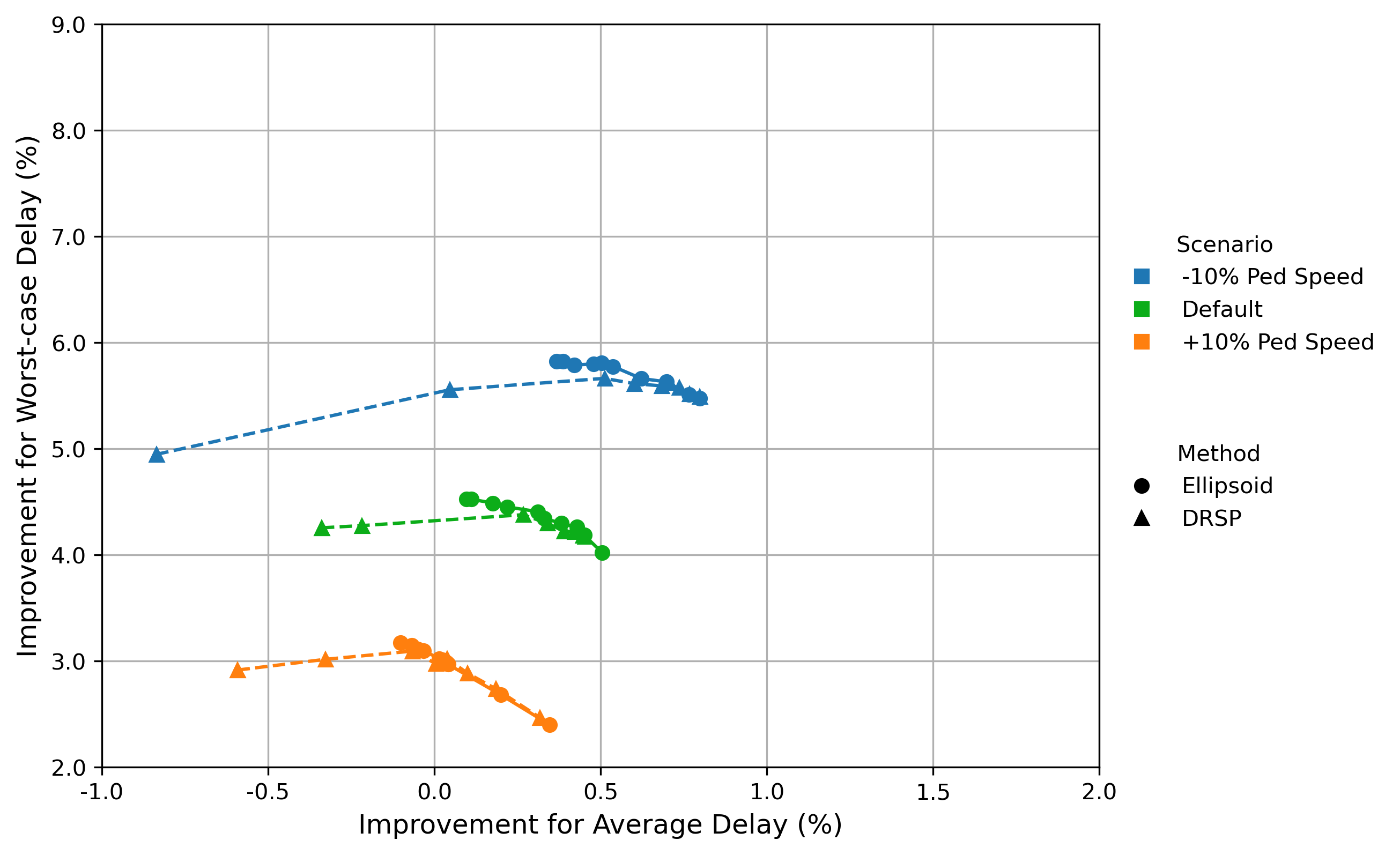}
        \caption{Pedestrian speed}
        \label{fig:12a}
    \end{subfigure}

    \begin{subfigure}{0.6\textwidth}
        \centering
        \includegraphics[width=\linewidth]{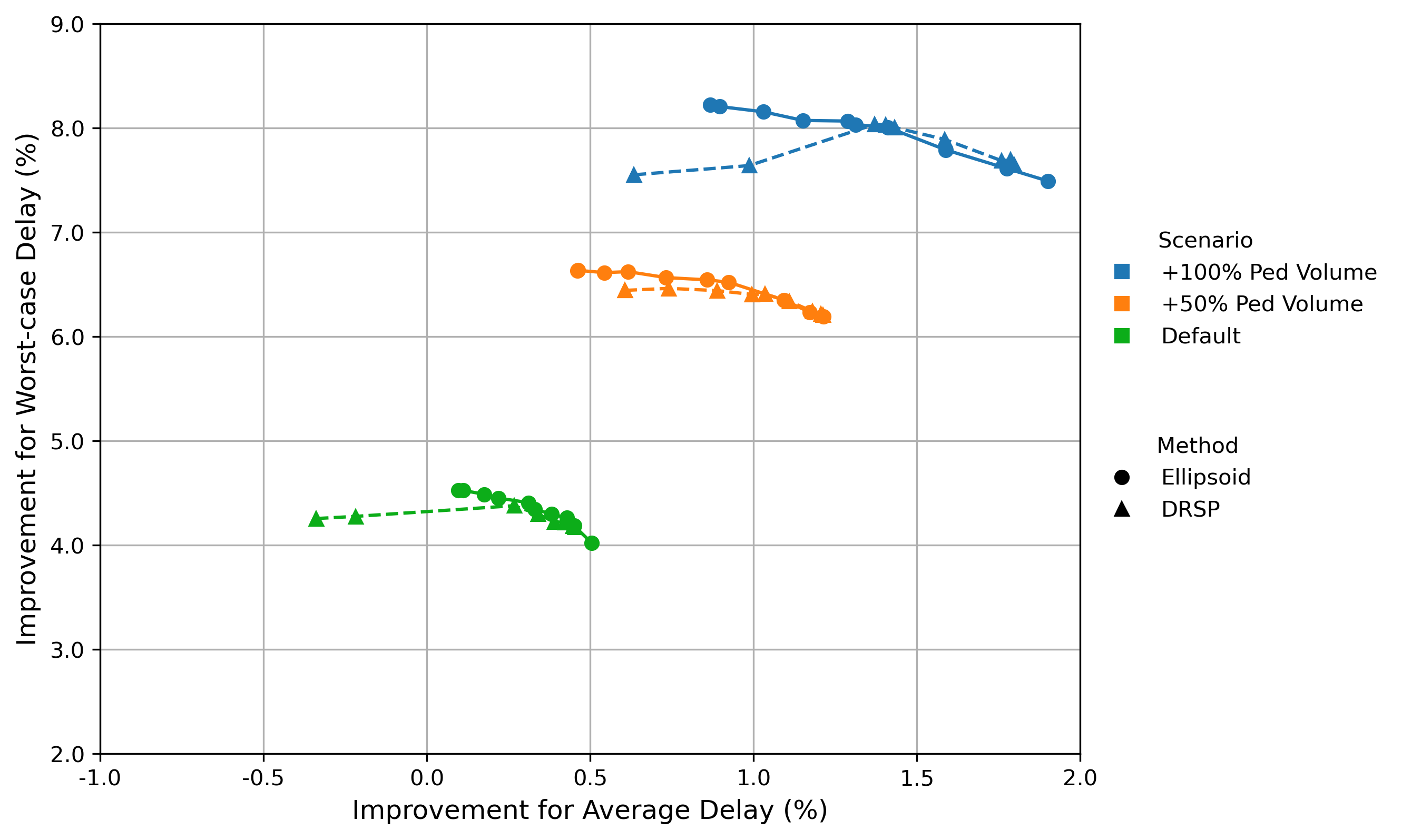}
        \caption{Pedestrian volume}
        \label{fig:12b}
    \end{subfigure}
    \caption{Performance improvement of selected robust methods under various environmental factors}
    \label{fig:12}
\end{figure}

Environmental conditions such as weather and pedestrian density can significantly affect sidewalk robot navigation. Weather conditions significantly influence pedestrian speed on sidewalks \citep{su16114813} due to reduced visibility, balance challenges, and surface conditions. For example, rain, wind, or snow typically slow movement, while light rain or extreme heat may incentivize pedestrians to walk faster to minimize exposure. To simulate these irregular weather conditions, pedestrian speed is adjusted by ±10\%, based on empirical evidence from \citet{FOSSUM2021102934,LIANG2020106811}. While VISWALK does not allow direct simulation of weather, its effects can be indirectly represented by adjusting relevant parameters. In addition, while the original set of experiments realistically accounts for hourly and daily pedestrian volume variations, extraordinary circumstances such as special events, holidays, and tourist peaks can cause occasional surges in foot traffic. These conditions are simulated by uniformly increasing the pedestrian volumes across the network by 50\% and 100\%. Using the same methodology applied to the robot design features, the improvements for average travel delays and worst travel delays achieved by robust methods over the SP method are derived and presented in Figure~\ref{fig:12}.

The relative benefits of robust routing diminish with increased average pedestrian speed. Specifically, the maximum improvement in worst-case delay declines from 5.8\% under a 10\% reduction in pedestrian speed and to 3.2\% under a 10\% increase. Similarly, the improvement in average delay drops from 0.8\% to 0.35\% across the same range. In contrast, increased pedestrian volumes lead to greater gains from robust methods. When pedestrian volume is doubled, the maximum improvement in worst-case delay rises from 4.5\% in the default case to 8.2\%, while the gain in average delay rises from 0.5\% to 1.9\%.

These results highlight the robustness and reliability of the selected models, whose performance over the SP approach,  increases significantly even under moderate surges in pedestrian volumes and adverse weather conditions which reduce the average pedestrian speed. 

The two robust methods exhibit similar trends to those observed in  Section~\ref{sec: robot design features}. The Ellipsoidal method consistently outperforms the DRSP method, indicating that when reliability is prioritized over efficiency, the Ellipsoidal method offers superior performance across all scenarios. When the average delay is the primary focus, the two robust methods show comparable performance.

\subsection{Operational Insights}
\label{sec:operational_insights}
The analyses presented in this study provide several actionable insights for logistics operators deploying sidewalk delivery robots in urban settings. Robust routing approaches (particularly the Ellipsoidal and DRSP methods) will bring significant benefits in complex environments characterized by pedestrian congestion and variability. These gains can be amplified during weather events that reduce pedestrian speed. For operators managing time-sensitive services like on-demand food delivery and grocery delivery, robust routing approaches can help improve service reliability and customer satisfaction. These considerations easily apply to many scenarios sharing similar characteristics with Stockholm city center, such as compact urban form, high pedestrian volumes, and variable weather conditions.

To further quantify these benefits from an operational perspective, we examine the impact of robust routing on fleet requirements. We consider a simple tour where a robot departs from a depot, visits an OD pair, and returns to the depot. This tour is constructed from three consecutive OD pairs extracted from the set of 100 OD pairs analyzed in previous sections. For this tour, we compute travel times across all 404 validation scenarios for both the conventional shortest path and the Ellipsoidal robust routing with  \( \lambda =3\). To present more challenging navigation behavior, we use the conservative robot setting while keeping all other settings as default. We rank all scenarios by the travel time of the conventional shortest path and evaluate fleet requirements at multiple percentiles (10th, 20th, ..., worst case). For each percentile, we report the corresponding tour travel times and implied fleet sizes for both methods using a workload-based fleet-sizing approximation \(N= \left\lceil \frac{K\,T}{H} \right\rceil\), where \(H\) is the required delivery frequency and \(K\) is the number of simultaneous orders (see Appendix~\ref{app:fleet_size}). 

The results show that under typical conditions, the required fleet size is identical for both routing methods, whereas under increasingly adverse scenarios, robust routing increasingly leads to fewer robots than conventional shortest-path planning. In the worst-case scenario, for example, the robust method reduces the tour travel time from about 1227 s to 1149 s, i.e., roughly 6.4\%. With \(H=20 \, \text{min}\) and \(K=1\), this reduction decreases the required fleet size by one robot; when \(K=50\), the difference increases to four robots, which is consistent with the scale of the 6.4\% travel time reduction.

Repeating this analysis under other conditions, such as slower or wider robots, higher pedestrian volumes, or adverse weather conditions, yields even larger difference in required fleet size. These results indicate that, beyond improving reliability indicators, robust routing can also reduce the fleet size needed to sustain a given level of service under unfavorable conditions, which is directly relevant for logistics operators. 

The sensitivity analyses also suggest that certain robot designs inherently benefit more from robust routing. While several designs are currently being tested by manufacturers, it is unlikely that the maximum allowed speeds will be higher than pedestrian speeds (around 5 km/h) for safety reasons. It is also probable that manufacturers, at least in the early stages, will adopt more conservative navigation behavior. These characteristics increase travel time variability, reinforcing the operational value of robust routing approaches in real-world deployments.

From an efficiency standpoint, operators may consider hybrid strategies where robust route planning is activated only during periods characterized by higher uncertainty (e.g., peak hours, poor weather). This targeted deployment can reduce computational requirements while securing reliability when it matters most. Finally, practical aspects such as data availability and tuning effort should not be neglected in the specific method selection.

\section{Conclusion}
\label{Sec:Conclusion}

Sidewalk delivery robots represent a promising opportunity for enhancing last-mile distribution in urban settings. However, their efficiency in environments characterized by high pedestrian volumes and unforeseen obstacles is relatively unknown. This study addresses this gap by investigating route planning for sidewalk robots by explicitly formulating the problem as a robust route planning task and integrating it with the simulation of sidewalk pedestrian flows.

Four robust approaches are analyzed: two ``traditional" formulations (Budgeted and Ellipsoidal uncertainty), and two data-driven methods (the SVC-based method and DRSP approach). The application of SVC in robust optimization is particularly noteworthy, as it merges machine learning and optimization techniques to model uncertainty in high-dimensional, complex data settings. To address computational issues, we implement the TSC-DS algorithm and propose an extension using hierarchical clustering to generate more meaningful feature groupings. Pedestrian simulation is adopted to realistically account for travel time variability due to different uncertainties, including environmental and operational ones. 

A comparative analysis with the conventional SP approach shows that robust approaches consistently lead to more reliable and efficient routing decisions for the sidewalk. The majority of the analyzed 500 OD pairs exhibit significant variability, highlighting the importance of robust methods for the sidewalk delivery route planning. Of all the methods tested, the Ellipsoidal and well-tuned DRSP methods achieve the best overall performance. Although the SVC-based method offers high flexibility by avoiding prior assumptions on the uncertainty structure, it does not perform better than Ellipsoidal or DRSP methods likely due to limited complexity and size of the problem studied.  

Finally, we perform a systematic sensitivity analysis to assess the benefits of the two most efficient robust models (Ellipsoidal and DRSP) under various robot design and environmental conditions. The results further confirm the advantages of robust approaches, particularly for delivery robots with larger width, lower design speeds, and more conservative navigation behaviors. Moreover, these advantages are even greater in dense pedestrian settings and when average walking speeds decrease, underscoring the value of robust methods under challenging real-world conditions.

The analyses of this study shed light on different questions, while leaving others for future research. The simulated travel time matrix may exhibit regularities, making the distribution of travel times, and ultimately favoring standard methods like the ellipsoidal. The performance of the data-driven methods, including both SVC and DRSP, could improve with access to larger and more diverse training datasets. The assumption of full travel time observability in the simulation-based approach is also ideal since in real-world applications, trip data from sidewalk delivery robots is likely to be sparser and incomplete. Future research should therefore focus on methods capable of handling partial observability. Scalability challenges of robust routing methods arising with larger networks are another promising direction of research. Future work could integrate the proposed uncertainty-aware path costs into robust tour-planning, in which travel-time uncertainty may influence both route reliability and the optimal sequencing of customers. Additional case studies involving shared and multi-modal spaces would further support the implementation of robust robot-based last-mile delivery.

\section*{Acknowledgments}
The study was partly funded by Digital Futures under the ISMIR project, \textit{Investigating Sidewalks' Mobility and Improving it with Robots}. The authors thank Bartolomeo Stellato at Princeton University for insightful discussions on robust optimization and uncertainty modeling. They also gratefully acknowledge PTV Group for providing access to a VISWALK license for this research.

\clearpage 
\appendix
\section{Glossary of acronyms, methods, and key parameters used in the paper}
\label{app:glossary}

\begin{table}[htbp]
\centering
\caption{Glossary of acronyms, methods, and key parameters used in the paper}
\label{tab:glossary}
\renewcommand{\arraystretch}{1.15}
\setlength{\tabcolsep}{6pt}
\begin{tabular}{p{4cm} p{10cm}}
\hline
\textbf{Term} & \textbf{Description} \\
\hline
SP & Shortest Path problem with deterministic travel times. \\
RSPP & Robust Shortest Path Problem, minimizing worst-case travel time over an uncertainty set. \\
DRSP & Distributionally Robust Shortest Path Problem, optimizing against worst-case distributions within an ambiguity set. \\
Budgeted uncertainty & Uncertainty model with deviation budget $\Gamma$.\\
Ellipsoidal uncertainty & Uncertainty modeled by an ellipsoid defined by mean and covariance of samples. \\
SVC uncertainty& Data-driven uncertainty modeled by Support Vector Clustering.\\
SKL / MKL & Single-Kernel Learning / Multiple-Kernel Learning variants of SVC. \\
TSC-DS & Two-Stage Clustering with Dimensional Separation to reduce dimensionality in SVC. \\
Wasserstein ambiguity set & Set of probability distributions within Wasserstein distance $\varepsilon_N$ from empirical distribution. \\
METT & Mean-Excess Travel Time, a tail-risk measure of travel time. \\
VISWALK & Microscopic pedestrian simulation software by PTV. \\
SFM & Social Force Model governing pedestrian and robot motion in simulation. \\
OD pair & Origin-Destination pair in the sidewalk network.\\
$\Gamma$ & Budget parameter controlling conservatism in budgeted uncertainty. \\
$\lambda$ & Size parameter of ellipsoidal uncertainty set. \\
$\nu$ & Regularization parameter in SVC controlling fraction of support vectors. \\
$\varepsilon_N$ & Radius of Wasserstein ambiguity set in DRSP. \\
$\alpha$ & Tail probability in METT risk measure. \\
\hline
\end{tabular}
\end{table}

\section{Derivation of RSP model based on SVC}
\label{app:svc-derivation}

This appendix provides the detailed derivation of the RSP model based on SVC, which is presented in Section~\ref{sec:RRPA}. The SKL-based SVC with the WGIK was proposed by \citet{SHANG2017464}. It utilizes SVs to define boundaries in feature space, grouping data points into clusters by finding the smallest sphere that encloses them. This approach not only manages correlated uncertainties, resulting in asymmetric uncertainty sets, but it also features adaptive complexity, embodying a nonparametric approach. The convex polyhedral uncertainty set generated can ensure the robust counterpart problem of the same type as the deterministic problem. 

The SKL-based SVC model uses a nonlinear mapping \(\phi(u)\) to project observations into high-dimensional feature space, and then finds the smallest hyperplane that encloses all data. The mapping function \(\phi(u)\) is complicated but it can be calculated indirectly by evaluating kernel function \(K(\mathbf{u}^i,\mathbf{u}^j)=\phi(\mathbf{u}^i)^T\phi(\mathbf{u}^j)\). The dual formulation of SVC model is given below:
    \begin{equation}\label{eq:A.1}
    \begin{aligned}
    \underset{\mathbf{\alpha}}{\min} 
    \quad & 
    \sum_{i=1}^N \sum_{j=1}^N \alpha_i \alpha_j K(\mathbf{u}^i,\mathbf{u}^j) - \sum_{i=1}^N \alpha_iK(\mathbf{u}^i,\mathbf{u}^i)
    \\ \text{s.t.} \quad &  0 \leq \alpha_i \leq 1/Nv, i = 1,...,N
    \\ \quad &   \sum_{i=1}^N \alpha_i =1
    \end{aligned}
    \end{equation}
    A regularization parameter \(v \in (0,1]\) is introduced to control the conservatism degree of the uncertainty set. It is an upper bound on the fraction of outliers and a lower bound on the fraction of support vectors \citep{SHANG2017464} . The kernel function WGIK is calculated in Eq. \eqref{eq:A.2}:
    \begin{equation}\label{eq:A.2}
    K(\mathbf{u},\mathbf{v})=\sum_{k=1}^n l_k -||\mathbf{Q}(\mathbf{u}-\mathbf{v})||_1
    \end{equation}
    where \(\mathbf{Q}\) is a weighting matrix generated from covariance matrix  \(\mathbf{Q}=\mathbf{\Sigma}^{-\frac{1}{2}}\). The unbiased estimation of the covariance matrix \(\mathbf{\Sigma}\) can be obtained in Eq.\eqref{eq:A.3}
    \begin{equation}\label{eq:A.3}
    \mathbf{\Sigma}=\frac{1}{N-1}\left[\sum_{k=1}^n \mathbf{u}^i(\mathbf{u}^i)^T - \left(\sum_{k=1}^n\mathbf{u}^i\right)\left(\sum_{k=1}^n\mathbf{u}^i\right)^T\right]
    \end{equation}
    The kernel parameters \(l_k\) should satisfy Eq.\eqref{eq:A.4}, then the kernel matrix \(\mathbf{K}\) can be positive definite. Note that if Eq.\eqref{eq:A.4} is satisfied, no matter what the value of \(l_k\) is, it will not affect the induced SVC model. 
    \begin{equation}\label{eq:A.4}
    l_k > \underset{1 \leq i \leq N}{\max} \mathbf{q}_k^T\mathbf{u}^i-\underset{1 \leq i \leq N}{\min} \mathbf{q}_k^T\mathbf{u}^i 
    \end{equation}
    
    After solving the model \eqref{eq:A.1} with kernel function WGIK, the index sets of all support vectors and boundary support vectors can be defined according to the following Eqs. \eqref{eq:A.5}\eqref{eq:A.6}
    \begin{equation}\label{eq:A.5}
    SV=\{i|\alpha_i > 0, \forall i\}
    \end{equation}
    \begin{equation}\label{eq:A.6}
    BSV=\{i|0< \alpha_i < 1/Nv, \forall i\}
    \end{equation}
    Besides, an explicit expression of the data-driven uncertainty set can be derived as a set of linear inequalities:
    \begin{equation}\label{eq:A.7}
    U^{SVC}=  \left\{ \mathbf{u} \middle| 
    \begin{aligned}
        & \exists \mathbf{v}_i, i \in SV  s.t.
    \\  & \sum_{i \in SV} \alpha_i \cdot \mathbf{1}^T \mathbf{v}_i \leq \theta
    \\  & -\mathbf{v}_i \leq \mathbf{Q}(\mathbf{u}-\mathbf{u}^i) \leq \mathbf{v}_i, \forall i \in SV
    \end{aligned}   
    \right\} 
    \end{equation}
    where \(\theta= \sum_{i \in SV} \alpha_i ||\mathbf{Q}(\mathbf{u}^{i'}-\mathbf{u}^i)||_1\), \(i'\) can be any index in \(BSV\), and \(\mathbf{v_i}\) are auxiliary variables introduced.
    
    In order to apply the above SVC-based uncertainty set into robust shortest path problem, Eq.\eqref{eq:2} can be simplified by moving the inner objective function into a constraint
    \begin{equation}\label{eq:A.8}
    \begin{aligned}
    \underset{\mathbf{x}\in X, b}{\min} \quad & b \\
    \text{s.t.} \quad & \underset{\mathbf{u}\in U}{\max} \mathbf{u}^T \mathbf{x} \leq b  
    \end{aligned}
    \end{equation}
    The left hand side (LHS) of constraint in Eq. \eqref{eq:A.8} can be reformulated into the linear programming model Eq.\eqref{eq:A.9} by replacing the uncertainty set as linear inequalities constraints in Eq.\eqref{eq:A.7}. 
    \begin{equation}\label{eq:A.9}
    \begin{aligned}
    \underset{\mathbf{u},\mathbf{v_i}}{\max} 
    \quad & 
    \mathbf{u}^T \mathbf{x} \\
    \text{s.t.} \quad &  \sum_{i \in SV} \alpha_i \cdot \mathbf{1}^T \mathbf{v}_i \leq \theta\\
    \quad &   -\mathbf{v}_i \leq \mathbf{Q}(\mathbf{u}-\mathbf{u}^i) \leq \mathbf{v}_i, \forall i \in SV
    \end{aligned}
    \end{equation}
    By introducing Lagrange multipliers \(\mathbf{\lambda_i}\), \(\mathbf{\mu_i}\), and \(\eta\), the model \eqref{eq:A.9} can be translated into its dual form: 
    \begin{equation}\label{eq:A.10}
    \begin{aligned}
    \underset{\mathbf{\lambda_i},\mathbf{\mu_i},\mathbf{\eta}}{\min} \quad & 
    \sum_{i \in SV} (\mathbf{\mu_i} - \mathbf{\lambda_i})^T \mathbf{Q}\mathbf{u}^i + \eta\theta \\
    \text{s.t.} \quad &  
    \sum_{i \in SV} \mathbf{Q}(\mathbf{\lambda_i} - \mathbf{\mu_i}) + \mathbf{x}  = \mathbf{0}
    \\ \quad &   \mathbf{\lambda_i} + \mathbf{\mu_i} = \eta \cdot \alpha_i \cdot \mathbf{1}, \mathbf{\lambda_i},\mathbf{\mu_i} \in \mathbb{R}_+^n, \forall i \in SV
    \\ \quad & \eta \geq 0
    \end{aligned}
    \end{equation}
    We know that, for any feasible \(\mathbf{\lambda_i}\), \(\mathbf{\mu_i}\), \(\eta\), and optimal solution \(\mathbf{u}\), \(\mathbf{v}_i\) to the primal problem \eqref{eq:A.9}, there always exists \(\sum_{i \in SV} (\mathbf{\mu_i} - \mathbf{\lambda_i})^T \mathbf{Q}\mathbf{u}^i + \eta\theta \geq \mathbf{u}^T \mathbf{x}\). Moreover, according to \citet{SHANG2017464}, the objective values of the primal and dual coincide due to strong duality of LP. Thus, the Eq.\eqref{eq:A.8} can be written as Eq.\eqref{eq:A.11}:
    \begin{equation}\label{eq:A.11}
    \begin{aligned}
    \underset{\mathbf{x},\mathbf{\lambda_i},\mathbf{\mu_i},\mathbf{\eta}}{\min} \quad & b
    \\ \text{s.t.} \quad &  
    \sum_{i \in SV} (\mathbf{\mu_i} - \mathbf{\lambda_i})^T \mathbf{Q}\mathbf{u}^i + \eta\theta  \leq b
    \\ \quad & \sum_{i \in SV} \mathbf{Q}(\mathbf{\lambda_i} - \mathbf{\mu_i}) + \mathbf{x}  = \mathbf{0}
    \\ \quad &   \mathbf{\lambda_i} + \mathbf{\mu_i} = \eta \cdot \alpha_i \cdot \mathbf{1}, \mathbf{\lambda_i},\mathbf{\mu_i} \in \mathbb{R}_+^n, \forall i \in SV
    \\ \quad & \eta \geq 0
    \\ \quad & \mathbf{x}\in X
    \end{aligned}
    \end{equation}
    Substituting \(X\) with constraints for shortest path problem, the final model for robust shortest path problem with WGIK-SVC-based uncertainty set can be defined as follows:
    \begin{equation}\label{eq:A.12}
    \begin{aligned}
    \underset{x_{pq},\mathbf{\lambda_i},\mathbf{\mu_i},\mathbf{\eta}}{\min} \quad & b 
\\ \text{s.t.} \quad &  
\sum_{i \in SV} (\mathbf{\mu_i} - \mathbf{\lambda_i})^T \mathbf{Q}\mathbf{u}^i + \eta\theta \leq b 
\\ \quad & \sum_{i \in SV} \mathbf{Q}(\mathbf{\lambda_i} - \mathbf{\mu_i}) + \mathbf{x}  =  \mathbf{0} 
\\ \quad &   \mathbf{\lambda_i} + \mathbf{\mu_i} = \eta \cdot \alpha_i \cdot \mathbf{1},  \forall i \in SV 
\\ \quad & \sum_{\{q:(p,q) \in A\}} x_{pq} - \sum_{\{q:(q,p) \in A\}} x_{qp}=
\begin{cases}
    1 & \text{if } p=s \\
    -1 & \text{if } q=t \\
    0 & \text{otherwise }
\end{cases} 
\\ \quad & \mathbf{\lambda_i},\mathbf{\mu_i}\in \mathbb{R}_+^n, \eta \geq 0, x_{pq} \in \{0,1\} 
    \end{aligned}
    \end{equation}

\section{Full Kinematic Sensitivity Analysis Results for SFM Parameter Combinations}
\label{app:sfm_full_results}

Table~\ref{tab:sfm_full_results} reports the full set of kinematic performance indicators
for all tested combinations of SFM parameters. For reference, the VISWALK default pedestrian parameter set is also included as a baseline (bold, first row). For each parameter set, results are averaged over 10 replications with identical scenario settings and different random seeds. The reported indicators include the 5th percentile of implied turning radius, 95th percentile of absolute yaw rate, 95th percentile of absolute lateral acceleration, mean travel time, and mean speed.

\begin{longtable}{ccccSSSSS}
\caption{Full kinematic sensitivity analysis results for all SFM parameter combinations.}
\label{tab:sfm_full_results} \\
\toprule
$\tau$ &
$\lambda$ &
$A_{\text{soc}}^{\text{iso}}$ &
$B_{\text{soc}}^{\text{iso}}$ &
\multicolumn{1}{c}{\makecell{Turn radius $p_5$ \\ (m)}} &
\multicolumn{1}{c}{\makecell{Yaw rate $p_{95}$ \\ (rad/s)}} &
\multicolumn{1}{c}{\makecell{Lat. acc. $p_{95}$ \\ (m/s\textsuperscript{2})}} &
\multicolumn{1}{c}{\makecell{Travel time \\ (s)}} &
\multicolumn{1}{c}{\makecell{Speed \\ (m/s)}} \\
\midrule
\endfirsthead

\toprule
$\tau$ &
$\lambda$ &
$A_{\text{soc}}^{\text{iso}}$ &
$B_{\text{soc}}^{\text{iso}}$ &
\multicolumn{1}{c}{\makecell{Turn radius $p_5$ \\ (m)}} &
\multicolumn{1}{c}{\makecell{Yaw rate $p_{95}$ \\ (rad/s)}} &
\multicolumn{1}{c}{\makecell{Lat. acc. $p_{95}$ \\ (m/s\textsuperscript{2})}} &
\multicolumn{1}{c}{\makecell{Travel time \\ (s)}} &
\multicolumn{1}{c}{\makecell{Speed \\ (m/s)}} \\
\midrule
\endhead

\midrule
\multicolumn{9}{r}{\textit{Continued on next page}} \\
\endfoot

\bottomrule
\endlastfoot

\textbf{0.4} & \textbf{0.176}  & \textbf{2.72} & \textbf{0.2}  & \textbf{1.954} & \textbf{0.571} & \textbf{0.689} & \textbf{61.13} & \textbf{1.266} \\
0.4 & 0.2  & 1 & 0.2  & 2.857 & 0.421 & 0.517 & 60.45 & 1.278 \\
0.4 & 0.2  & 1 & 0.35 & 2.73  & 0.418 & 0.502 & 61.78 & 1.253 \\
0.4 & 0.2  & 1 & 0.5  & 2.655 & 0.427 & 0.502 & 62.77 & 1.235 \\
0.4 & 0.2  & 2 & 0.2  & 2.04  & 0.545 & 0.652 & 61.89 & 1.251 \\
0.4 & 0.2  & 2 & 0.35 & 2.003 & 0.553 & 0.663 & 62.61 & 1.239 \\
0.4 & 0.2  & 2 & 0.5  & 1.73  & 0.611 & 0.687 & 65.02 & 1.197 \\
0.4 & 0.2  & 3 & 0.2  & 1.703 & 0.672 & 0.793 & 62.1  & 1.249 \\
0.4 & 0.2  & 3 & 0.35 & 1.315 & 0.795 & 0.865 & 63.66 & 1.222 \\
0.4 & 0.2  & 3 & 0.5  & 1.221 & 0.822 & 0.853 & 67.46 & 1.161 \\
0.4 & 0.45 & 1 & 0.2  & 2.921 & 0.406 & 0.507 & 59.88 & 1.288 \\
0.4 & 0.45 & 1 & 0.35 & 2.693 & 0.436 & 0.528 & 61.94 & 1.251 \\
0.4 & 0.45 & 1 & 0.5  & 2.82  & 0.411 & 0.486 & 61.64 & 1.256 \\
0.4 & 0.45 & 2 & 0.2  & 2.059 & 0.561 & 0.658 & 61.34 & 1.263 \\
0.4 & 0.45 & 2 & 0.35 & 1.851 & 0.63  & 0.739 & 62.42 & 1.244 \\
0.4 & 0.45 & 2 & 0.5  & 1.852 & 0.606 & 0.711 & 63.24 & 1.23  \\
0.4 & 0.45 & 3 & 0.2  & 1.589 & 0.788 & 0.916 & 62.16 & 1.25  \\
0.4 & 0.45 & 3 & 0.35 & 1.407 & 0.782 & 0.844 & 65.57 & 1.193 \\
0.4 & 0.45 & 3 & 0.5  & 1.186 & 0.84  & 0.899 & 66.54 & 1.177 \\
0.4 & 0.7  & 1 & 0.2  & 2.993 & 0.39  & 0.488 & 60.01 & 1.286 \\
0.4 & 0.7  & 1 & 0.35 & 2.921 & 0.406 & 0.492 & 60.63 & 1.276 \\
0.4 & 0.7  & 1 & 0.5  & 2.795 & 0.416 & 0.499 & 61.92 & 1.252 \\
0.4 & 0.7  & 2 & 0.2  & 2.024 & 0.572 & 0.691 & 62.18 & 1.248 \\
0.4 & 0.7  & 2 & 0.35 & 1.949 & 0.603 & 0.709 & 62.96 & 1.236 \\
0.4 & 0.7  & 2 & 0.5  & 1.753 & 0.61  & 0.729 & 62.8  & 1.238 \\
0.4 & 0.7  & 3 & 0.2  & 1.445 & 0.83  & 0.961 & 61.9  & 1.255 \\
0.4 & 0.7  & 3 & 0.35 & 1.391 & 0.781 & 0.851 & 64.74 & 1.207 \\
0.4 & 0.7  & 3 & 0.5  & 1.16  & 0.889 & 0.966 & 65.9  & 1.194 \\
0.8 & 0.2  & 1 & 0.2  & 2.628 & 0.429 & 0.507 & 62.43 & 1.25  \\
0.8 & 0.2  & 1 & 0.35 & 2.379 & 0.465 & 0.537 & 63.41 & 1.231 \\
0.8 & 0.2  & 1 & 0.5  & 1.79  & 0.587 & 0.595 & 66.57 & 1.183 \\
0.8 & 0.2  & 2 & 0.2  & 1.952 & 0.558 & 0.645 & 62.5  & 1.248 \\
0.8 & 0.2  & 2 & 0.35 & 1.311 & 0.72  & 0.722 & 67.5  & 1.172 \\
0.8 & 0.2  & 2 & 0.5  & 1.081 & 0.823 & 0.766 & 69.12 & 1.144 \\
0.8 & 0.2  & 3 & 0.2  & 1.413 & 0.765 & 0.757 & 65.01 & 1.21  \\
0.8 & 0.2  & 3 & 0.35 & 0.915 & 1.037 & 0.931 & 68.89 & 1.152 \\
0.8 & 0.2  & 3 & 0.5  & 0.771 & 1.062 & 0.933 & 71.32 & 1.119 \\
0.8 & 0.45 & 1 & 0.2  & 2.659 & 0.435 & 0.508 & 62.39 & 1.25  \\
0.8 & 0.45 & 1 & 0.35 & 2.313 & 0.464 & 0.528 & 63.25 & 1.233 \\
0.8 & 0.45 & 1 & 0.5  & 1.74  & 0.578 & 0.599 & 66.34 & 1.186 \\
0.8 & 0.45 & 2 & 0.2  & 1.947 & 0.566 & 0.652 & 62.5  & 1.248 \\
0.8 & 0.45 & 2 & 0.35 & 1.167 & 0.728 & 0.709 & 67.46 & 1.173 \\
0.8 & 0.45 & 2 & 0.5  & 0.973 & 0.875 & 0.8   & 69.53 & 1.14  \\
0.8 & 0.45 & 3 & 0.2  & 1.533 & 0.74  & 0.756 & 64.69 & 1.216 \\
0.8 & 0.45 & 3 & 0.35 & 0.759 & 1.095 & 0.97  & 69.03 & 1.152 \\
0.8 & 0.45 & 3 & 0.5  & 0.736 & 1.139 & 0.951 & 71.54 & 1.118 \\
0.8 & 0.7  & 1 & 0.2  & 2.601 & 0.44  & 0.533 & 62.39 & 1.251 \\
0.8 & 0.7  & 1 & 0.35 & 2.324 & 0.47  & 0.537 & 63.21 & 1.235 \\
0.8 & 0.7  & 1 & 0.5  & 1.881 & 0.587 & 0.599 & 66.8  & 1.185 \\
0.8 & 0.7  & 2 & 0.2  & 1.8   & 0.572 & 0.658 & 62.48 & 1.249 \\
0.8 & 0.7  & 2 & 0.35 & 1.382 & 0.72  & 0.725 & 66.11 & 1.198 \\
0.8 & 0.7  & 2 & 0.5  & 0.976 & 0.875 & 0.808 & 69.37 & 1.144 \\
0.8 & 0.7  & 3 & 0.2  & 1.451 & 0.776 & 0.777 & 64.25 & 1.222 \\
0.8 & 0.7  & 3 & 0.35 & 0.736 & 1.176 & 1.02  & 68.83 & 1.155 \\
0.8 & 0.7  & 3 & 0.5  & 0.656 & 1.275 & 1.025 & 75.6  & 1.082 \\
1.2 & 0.2  & 1 & 0.2  & 1.556 & 0.587 & 0.543 & 65.67 & 1.198 \\
1.2 & 0.2  & 1 & 0.35 & 1.265 & 0.677 & 0.586 & 67.11 & 1.177 \\
1.2 & 0.2  & 1 & 0.5  & 1.104 & 0.845 & 0.605 & 69.37 & 1.146 \\
1.2 & 0.2  & 2 & 0.2  & 1.28  & 0.713 & 0.593 & 69.42 & 1.14  \\
1.2 & 0.2  & 2 & 0.35 & 0.856 & 1.004 & 0.753 & 72.2  & 1.101 \\
1.2 & 0.2  & 2 & 0.5  & 0.72  & 1.2   & 0.781 & 75.27 & 1.069 \\
1.2 & 0.2  & 3 & 0.2  & 1.075 & 0.833 & 0.67  & 70.0  & 1.133 \\
1.2 & 0.2  & 3 & 0.35 & 0.619 & 1.368 & 0.919 & 72.89 & 1.098 \\
1.2 & 0.2  & 3 & 0.5  & 0.487 & 1.476 & 1.012 & 79.52 & 1.023 \\
1.2 & 0.45 & 1 & 0.2  & 1.517 & 0.607 & 0.552 & 66.54 & 1.184 \\
1.2 & 0.45 & 1 & 0.35 & 1.204 & 0.683 & 0.589 & 68.23 & 1.161 \\
1.2 & 0.45 & 1 & 0.5  & 0.973 & 0.96  & 0.648 & 71.14 & 1.116 \\
1.2 & 0.45 & 2 & 0.2  & 1.212 & 0.721 & 0.595 & 69.87 & 1.136 \\
1.2 & 0.45 & 2 & 0.35 & 0.852 & 1.05  & 0.754 & 71.34 & 1.115 \\
1.2 & 0.45 & 2 & 0.5  & 0.699 & 1.187 & 0.811 & 74.49 & 1.075 \\
1.2 & 0.45 & 3 & 0.2  & 0.969 & 0.88  & 0.748 & 70.51 & 1.131 \\
1.2 & 0.45 & 3 & 0.35 & 0.709 & 1.159 & 0.871 & 72.54 & 1.104 \\
1.2 & 0.45 & 3 & 0.5  & 0.449 & 1.668 & 1.059 & 80.09 & 1.029 \\
1.2 & 0.7  & 1 & 0.2  & 1.458 & 0.623 & 0.549 & 66.54 & 1.185 \\
1.2 & 0.7  & 1 & 0.35 & 1.316 & 0.672 & 0.58  & 69.82 & 1.133 \\
1.2 & 0.7  & 1 & 0.5  & 1.095 & 0.854 & 0.623 & 70.73 & 1.121 \\
1.2 & 0.7  & 2 & 0.2  & 1.193 & 0.721 & 0.622 & 69.91 & 1.136 \\
1.2 & 0.7  & 2 & 0.35 & 0.85  & 0.949 & 0.762 & 71.14 & 1.119 \\
1.2 & 0.7  & 2 & 0.5  & 0.634 & 1.234 & 0.843 & 75.85 & 1.054 \\
1.2 & 0.7  & 3 & 0.2  & 0.992 & 0.891 & 0.731 & 71.01 & 1.122 \\
1.2 & 0.7  & 3 & 0.35 & 0.648 & 1.33  & 0.874 & 72.81 & 1.103 \\
1.2 & 0.7  & 3 & 0.5  & 0.42  & 1.704 & 1.098 & 78.51 & 1.048 \\


\end{longtable}

\section{Fleet size implications under different scenarios}
\label{app:fleet_size}

This appendix reports the fleet requirements of the conventional shortest path and the Ellipsoidal robust routing with \( \lambda =3\) at different scenarios. The higher percentiles correspond to more adverse scenarios. The fleet size is computed using the workload-based approximation \(N=\lceil K T / H \rceil\) with \(H=20\) min and \(K=1\), as well as \(H=20\) min and \(K=50\).

\begin{table}[htbp]
\centering
\caption{Fleet requirements of the conventional shortest path and the Ellipsoidal robust routing.}
\label{tab:fleet_size_percentiles}
\renewcommand{\arraystretch}{1.15}
\setlength{\tabcolsep}{6pt}
\begin{tabular}{c c c c c c c}
\hline
Percentile &
\makecell{SP\_TT \\ (s)}&
\makecell{Ellipsoid\_TT \\ (s)}&
\makecell{SP\_fleet\_size \\ (H=20,K=1)} &
\makecell{Ellipsoid\_fleet\_size \\ (H=20,K=1)} &
\makecell{SP\_fleet\_size \\ (H=20,K=50)} &
\makecell{Ellipsoid\_fleet\_size \\ (H=20,K=50)} \\

\hline
10  & 1089.71 & 1089.72 & 1.0 & 1.0 & 46.0 & 46.0 \\
20  & 1100.93 & 1090.12 & 1.0 & 1.0 & 46.0 & 46.0 \\
30  & 1112.04 & 1096.52 & 1.0 & 1.0 & 47.0 & 46.0 \\
40  & 1117.65 & 1106.42 & 1.0 & 1.0 & 47.0 & 47.0 \\
50  & 1125.36 & 1103.88 & 1.0 & 1.0 & 47.0 & 46.0 \\
60  & 1135.36 & 1104.48 & 1.0 & 1.0 & 48.0 & 47.0 \\
70  & 1147.57 & 1108.58 & 1.0 & 1.0 & 48.0 & 47.0 \\
80  & 1160.0  & 1106.2  & 1.0 & 1.0 & 49.0 & 47.0 \\
90  & 1179.1  & 1130.12 & 1.0 & 1.0 & 50.0 & 48.0 \\
100 & 1226.54 & 1148.68 & 2.0 & 1.0 & 52.0 & 48.0 \\
\hline
\end{tabular}
\end{table}

\clearpage
\bibliographystyle{apalike}
\bibliography{references}

\end{document}